\documentclass[11pt,twoside,final]{article}

\usepackage{hyperref}
\hypersetup{colorlinks,linkcolor={blue},citecolor={blue},urlcolor={red}} 

\usepackage{fullpage}

\usepackage{epsf}
\usepackage{fancyheadings}
\usepackage{graphics}
\usepackage{graphicx}
\usepackage{enumitem}
\usepackage{float}

\usepackage{color}

\usepackage{amsthm}
\usepackage{amsfonts}
\usepackage{amsmath}
\usepackage{amssymb}

\usepackage{algorithm}
\usepackage{mathabx}
\usepackage{algorithmic}
\usepackage{bbm}

\usepackage{caption}
\usepackage{subcaption}

\usepackage[square, numbers]{natbib}

\theoremstyle{plain}

\newtheorem{theorem}{Theorem}

\newtheorem{lemma}{Lemma}

\theoremstyle{definition}
\newtheorem{example}{Example}

\newlength{\widebarargwidth}
\newlength{\widebarargheight}
\newlength{\widebarargdepth}
\DeclareRobustCommand{\widebar}[1]{%
  \settowidth{\widebarargwidth}{\ensuremath{#1}}%
  \settoheight{\widebarargheight}{\ensuremath{#1}}%
  \settodepth{\widebarargdepth}{\ensuremath{#1}}%
  \addtolength{\widebarargwidth}{-0.3\widebarargheight}%
  \addtolength{\widebarargwidth}{-0.3\widebarargdepth}%
  \makebox[0pt][l]{\hspace{0.3\widebarargheight}%
    \hspace{0.3\widebarargdepth}%
    \addtolength{\widebarargheight}{0.3ex}%
    \rule[\widebarargheight]{0.95\widebarargwidth}{0.1ex}}%
  {#1}}

\makeatletter
\long\def\@makecaption#1#2{
        \vskip 0.8ex
        \setbox\@tempboxa\hbox{\small {\bf #1:} #2}
        \parindent 1.5em  
        \dimen0=\hsize
        \advance\dimen0 by -3em
        \ifdim \wd\@tempboxa >\dimen0
                \hbox to \hsize{
                        \parindent 0em
                        \hfil 
                        \parbox{\dimen0}{\def\baselinestretch{0.96}\small
                                {\bf #1.} #2
                                } 
                        \hfil}
        \else \hbox to \hsize{\hfil \box\@tempboxa \hfil}
        \fi
        }
\makeatother

\newcounter{relctr} 
\everydisplay\expandafter{\the\everydisplay\setcounter{relctr}{0}} 

\newcommand\labelrel[2]{%
  \begingroup
    \refstepcounter{relctr}%
    \stackrel{\textnormal{(\alph{relctr})}}{\mathstrut{#1}}%
    \originallabel{#2}%
  \endgroup
}
\AtBeginDocument{\let\originallabel\label} 

\long\def\comment#1{}

\newcommand{\Op}{\ensuremath{\mathbf{T}}}
\newcommand{\NoisyOp}{\ensuremath{\widehat{\mathbf{T}}}}
\newcommand{\contractPar}{\gamma}
\newcommand{\ele}{\theta}
\newcommand{\elestar}{\ele^{\star}}

\newcommand{\numstates}{\ensuremath{D}}
\newcommand{\Prob}{\mathbb{P}}
\newcommand{\RR}{\mathbb{R}}
\newcommand{\TranMat}{\mathbf{P}}
\newcommand{\reward}{\ensuremath{r}}
\newcommand{\obsmat}{\ensuremath{\textbf{Z}}}

\newcommand{\elebar}{\overline{\ele}}

\newcommand{\stateset}{\ensuremath{\mathcal{X}}}
\newcommand{\actionset}{\ensuremath{\mathcal{U}}}
\newcommand{\StateSpace}{\stateset}
\newcommand{\ActionSpace}{\actionset}
\newcommand{\state}{\ensuremath{x}}
\newcommand{\action}{\ensuremath{u}}
\newcommand{\policy}{\ensuremath{\pi}}
\newcommand{\policystar}{\ensuremath{\policy^\star}}
\newcommand{\qvalues}{\ensuremath{\ele}}
\newcommand{\qvaluesstar}{\ensuremath{\ele^\star}}

\newcommand{\QL}{\textsc{{QL }}}

\newcommand{\VRQL}{\textsc{VR-QL}}
\newcommand{\epochs}{\ensuremath{M}}
\newcommand{\epochlength}{\ensuremath{K}}
\newcommand{\recentersize}{\ensuremath{N_m}}
\newcommand{\VROp}{\mathcal{V}}

\newcommand{\RecenterOp}{\ensuremath{\overline{\Op}}_{\recentersize}}
\newcommand{\recentersample}{\ensuremath{\mathcal{D}_m}}
\newcommand{\stepsize}{\ensuremath{\alpha}}
\newcommand{\epochsampleset}{\ensuremath{\mathcal{C}_m}}

\newcommand{\obsreward}{\ensuremath{R}}
\newcommand{\probmat}{\TranMat}

\newcommand{\widgraph}[2]{\includegraphics[keepaspectratio,width=#1]{#2}}

\newcommand{\Id}{\ensuremath{\mathbf{I}}}
 
\newcommand{\VecSpan}[1]{\ensuremath{\| #1\|_{\mathrm{span}}}}

\newcommand{\VRQLspace}{\VRQL$\;$}
\newcommand{\inftynorm}[1]{\ensuremath{\| #1 \|_{\infty}}}

\newcommand{\NumEpoch}{M}

\newcommand{\NoisyReward}{R}

\newcommand{\Qnumobs}{N}
\newcommand{\card}[1]{| #1 |}
\newcommand{\randmat}{\textbf{Y}}

\newcommand{\minimax}{\ensuremath{\mathfrak{M}}}
\newcommand{\probinstance}{\ensuremath{\mathcal{P}}}

\DeclareMathOperator{\var}{Var}

\newcommand{\sgPar}{\sigma}
\newcommand{\Exs}{\mathbb{E}}

\newcommand{\pardelta}{\delta}

\newcommand{\TranMatQ}{\mathbf{P}}
\newcommand{\probmatQ}{\probmat}
\newcommand{\Indicator}{\mathbbm{1}}

\newcommand{\infnorm}[1]{\| #1 \|_{\infty}}

\newcommand{\pistar}{\ensuremath{\pi^\star}}
\newcommand{\SpGap}{\ensuremath{\Delta}}

\newcommand{\Qstar}{\elestar}
\newcommand{\Q}{\ele}
\newcommand{\Ncal}{\ensuremath{\mathcal{N}}}
\newcommand{\abs}[1]{| #1 |}

\newcommand{\Qbar}{\widebar{\Q}}
\newcommand{\numobs}{N}
\newcommand{\epoch}{m}
\newcommand{\EpLen}{\ensuremath{\epochlength}}

\newcommand{\Nm}{\ensuremath{\numobs_\epoch}}

\newcommand{\LogDM}{\ensuremath{\log_4(8\numstates \epochs/\pardelta)}}

\newcommand{\Qhat}{\widehat{\Q}}
\newcommand{\pihat}{\widehat{\pi}}
\newcommand{\pibar}{\overline{\pi}}
\newcommand{\rewardtil}{\widetilde{\reward}}
\newcommand{\obsmathat}{\widehat{\obsmat}}
\newcommand{\shiftedOp}{\mathbf{J}}
\newcommand{\NoisyShiftedOp}{\widehat{\shiftedOp}}
\newcommand{\rewardhat}{\ensuremath{\widehat{\NoisyReward}}}
\newcommand{\LogDMsq}{\ensuremath{\log_4(8\numstates\epochs^2/\pardelta)}}

\newcommand{\PolicySet}{\ensuremath{\mathbf{\Pi}}}
\newcommand{\OptPolicySet}{\ensuremath{\mathbf{\Pi^\star}}}
\newcommand{\OptpolicySet}{\OptPolicySet}

\newcommand{\Nu}{\nu}

\newcommand{\instance}{\TranMatQ, \reward}
\newcommand{\Dim}{D}

\newcommand{\geqEle}{\succeq}
\newcommand{\leqEle}{\preceq}
\newcommand{\lampar}{\lambda}

\newcommand{\taupar}{\tau}
\newcommand{\Wmat}{\ensuremath{\mathbf{W}}}
\newcommand{\Vstar}{V^\star}
\newcommand{\lambdaInstance}{\TranMatQ_\lambda, \reward_\lambda}

\newcommand{\defn}{\ensuremath{: \, =}}
\newcommand{\CompTermInstance}{\max_{\pistar \in \OptpolicySet}
  \inftynorm{ \Nu(\pistar ; \TransMatQ, \reward, \contractPar) }}

\newcommand{\probdistr}{\probinstance}

\newcommand{\Alter}{\mathcal{S}}

\newcommand{\CompTerm}{\ensuremath{\rho}}
\newcommand{\CompTermEntry}{\ensuremath{\tilde{\CompTerm}}}
\newcommand{\AllCompTerm}{\ensuremath{\nu}}
\newcommand{\RewardTerm}{\ensuremath{\sigma}}
\newcommand{\SigTerm}{\ensuremath{\varphi}}
\newcommand{\InvTerm}{\mathbf{U}}
\newcommand{\InvTermInd}[2]{\mathbf{U}_{#1, #2}}
\newcommand{\distr}{\ensuremath{P}}
\newcommand{\TransInd}[2]{\Trans_{#2, #1}}
\newcommand{\PerTransInd}[2]{\bar{\Trans}_{#2, #1}}
\newcommand{\PerTrans}{\bar{\Trans}}
\newcommand{\Perpolicy}{\bar{\policy}}
\newcommand{\Perqvalues}{\bar{\qvalues}}
\newcommand{\Perreward}{\bar{\reward}}
\newcommand{\Pmat}{\mathbf{P}}
\newcommand{\discount}{\gamma}
\newcommand{\OptPiOne}{{\policy_{1}}}
\newcommand{\OptPi}{\policy^*}
\newcommand{\qestimate}{\hat{\qvalues}_\Qnumobs}
\newcommand{\PmatBar}{\PerTrans}
\newcommand{\Trans}{\TranMatQ}
\newcommand{\AltProb}{\mathcal{Q}}

\newcommand{\twonorm}[1]{\ensuremath{\| #1 \|_{2}}}
\newcommand{\TVnorm}[1]{\ensuremath{\| #1 \|_{\text{TV}}}}
\newcommand{\dhel}[2]{\ensuremath{d_{\text{Hel}}\left(#1, #2 \right)}}
\newcommand{\statesec}{\ensuremath{y}}

\newcommand{\sapair}{\ensuremath{z}}
\newcommand{\optsapair}{\ensuremath{\bar{\sapair}}}
\newcommand{\opnorm}[1]{\matsnorm{#1}{\infty \rightarrow \infty}}
\newcommand{\OptPiTwo}{{\policy_2}}

\newcommand{\epochstar}{\epoch^*}

\newcommand{\TauOpt}{\tau^*}

\newcommand{\NM}{\numobs_\epochs}

\newcommand{\TauMax}{\tau_{\max}}
\newcommand{\SpanTerm}{\kappa}
\newcommand{\HelDist}{d_{\text{Hel}}}

\usepackage{xcolor}

\newenvironment{varalgorithm}[1]
  {\algorithm\renewcommand{\thealgorithm}{#1}}
  {\endalgorithm}

\newcommand{\matsnorm}[2]{|\!|\!| #1 |\!|\!|_{{#2}}}
  
\newcommand{\matnorminfty}[1]{\matsnorm{#1}{\infty \rightarrow \infty}}

\newcommand{\TransMatQ}{\TranMatQ}
\newcommand{\nLBBeta}{\beta}

\newcommand{\UNQ}{UNQ}
\newcommand{\LIP}{LIP}

\newcommand{\betahat}{\widehat{\beta}}

\begin{document}

\begin{center}


  {\bf{\LARGE{Instance-optimality in optimal value estimation:\\
        Adaptivity
        via variance-reduced $Q$-learning}}}
\vspace*{.2in}

{\large{
\begin{tabular}{ccc}
 Koulik Khamaru$^{\dagger, \ast}$ \quad Eric Xia$^{\dagger, \ast}$, \\
 Martin J. Wainwright$^{\dagger, \star}$ \quad  Michael I. Jordan$^{\dagger, \star}$
\end{tabular}
}}

\vspace*{.2in}

\begin{tabular}{c}
Department of Statistics$^\dagger$, and \\
Department of Electrical Engineering and Computer Sciences$^\star$ \\
UC Berkeley,  Berkeley, CA  94720
\end{tabular}

\vspace*{.2in}

\footnotetext{$\ast$ denotes equal contribution to this work.}

\today

\vspace*{.2in}

\begin{abstract}
  Various algorithms in reinforcement learning exhibit dramatic
  variability in their convergence rates and ultimate accuracy as a
  function of the problem structure.  Such instance-specific behavior
  is not captured by existing global minimax bounds, which are
  worst-case in nature.  We analyze the problem of estimating optimal
  $Q$-value functions for a discounted Markov decision process with
  discrete states and actions and identify an instance-dependent
  functional that controls the difficulty of estimation in the
  $\ell_\infty$-norm.  Using a local minimax framework, we show that
  this functional arises in lower bounds on the accuracy on any
  estimation procedure.  In the other direction, we establish the
  sharpness of our lower bounds, up to factors logarithmic in the
  state and action spaces, by analyzing a variance-reduced version of
  $Q$-learning.  Our theory provides a precise way of distinguishing
  ``easy'' problems from ``hard'' ones in the context of $Q$-learning,
  as illustrated by an ensemble with a continuum of difficulty.
\end{abstract}

\end{center}


\section{Introduction}

The need for data-driven decision-making has fueled tremendous
interest in Markov decision processes and reinforcement learning (RL).
Indeed, such techniques have found use cases across a wide range of
application domains~\citep{tobin2017domain, JMLR:v17:15-522,
  silver2016alphago}.  An intriguing fact is that in many
applications, RL algorithms behave far better than the theoretical
bounds provided by worst-case analyses would suggest.  This gap
provides impetus for a more refined \emph{instance-specific} analysis,
one which highlights the properties of a given instance that render it
``easy'' or ``difficult.''

Instance-dependent analysis of RL algorithms has become of substantial
interest in recent years~\citep[see, e.g.,][]{simchowitz2019non,zanette2019tighter,zanette2019almost,maillard2014hard,pananjady2020instancedependent,khamaru2020PE}. By
now, we have a fairly refined understanding of instance-dependence for
policy evaluation problems, including work on temporal difference (TD)
algorithms under the $\ell_2$-norm~\citep{pmlr-v75-bhandari18a,
  pmlr-v84-lakshminarayanan18a,dalal2018finite}, as well as bounds for
the LSTD estimator in the
$\ell_\infty$-norm~\citep{pananjady2020instancedependent}.  A subset
of the current authors~\citep{khamaru2020PE} provided a sharper
instance-dependent $\ell_\infty$-bounds for a variance-reduced version
of the TD$(0)$ algorithm, and showed that this algorithm is optimal in
a local non-asymptotic minimax sense.

For TD and LSTD methods, the underlying structure is linear in
nature---in particular, it corresponds to solving a linear system---a
property which greatly facilitates the analysis.  In the current
paper we undertake a similar instance-dependent analysis in the more
challenging setting of $Q$-learning, for which the underlying updates
are non-linear.  Our main contributions are to identify a natural
functional of the problem instance and show that it controls the
fundamental difficulty of estimating optimal $Q$-value functions.  We
do so by establishing non-asymptotic lower bounds within a local
minimax framework and matching those bounds, up to logarithmic factors, by
analyzing a version of variance-reduced
$Q$-learning~\citep{sidford2018near, sidford2018variance,
  wainwright2019variancereduced}.

This work is done in the context of Markov decision processes (MDPs) with a
finite set of states $\stateset$ and a finite set of possible actions
$\actionset$. We proceed to provide some background and notation to be
able to introduce the functional that plays a central role in our
analysis, and describe our contributions in more detail.


\subsection{Some background}
\label{SecSomeBack}

In a Markov decision process, the state $\state$ evolves dynamically
in time under the influence of the actions.  More
precisely, there is a collection of probability transition kernels,
$\{\TranMatQ_\action(\cdot\mid\state) \mid (\state, \action) \in
\stateset \times \actionset\}$, where $\TranMatQ_\action(\state' \mid
\state)$ denotes the transition to the state $\state'$ 
when the action $\action$ is taken at the current state $\state$.  In
addition, an MDP is equipped with a reward function $r$ that maps every
state-action pair, $(\state, \action)$, to a real number
$\reward(\state, \action)$. The reward $\reward(\state, \action)$ is
the reward received upon performing an action $\action$ in the state
$\state$.  Overall, a given MDP is characterized by the problem pair
$(\TranMatQ, \reward)$, along with a discount factor $\contractPar \in
(0,1)$.

A deterministic policy $\policy$ is a mapping $\stateset \to
\actionset$: the quantity $\policy(\state) \in \actionset$ indicates
the action to be taken in the state $\state$.  The value of a policy
is defined by the expected sum of discounted rewards in an infinite
sample path.  For a given policy $\policy$ and discount factor
$\contractPar \in (0, 1)$, the $Q$-function is given by
\begin{align}
\qvalues^\policy(\state, \action) &\defn \Exs\left[\sum_{k=0}^\infty
  \gamma^k r(\state_k, \action_k) \mid \state_0 = \state, \, \action_0
  = \action \right], \qquad \text{where $\action_k = \policy(\state_k)$
  for all $k \geq 1$.}
\end{align}
When both the state space $\stateset$ and action space $\actionset$
are finite, the $Q$-function $\Q$ can be conveniently represented as
an element of $\RR^{\card{\stateset} \times \card{\actionset}}$.

There are various observation models in reinforcement learning, and in
this paper we study the \emph{generative setting} in which we have
the ability to draw next-state samples from the MDP when initialized
with an arbitrary state-action pair $(\state, \action)$.  More
precisely, we are given a collection of $\numobs$ i.i.d.\ samples of
the form $\{(\obsmat_k, \NoisyReward_k) \}_{k=1}^\numobs$, where both
$\obsmat_k$ and $\NoisyReward_k$ are random matrices in
$\RR^{\card{\stateset} \times \card{\actionset}}$.  For each
state-action pair $(\state, \action)$, the entry $\obsmat_k(\state,
\action)$ is drawn according to the transition kernel
$\TranMatQ_\action(\cdot \mid \state)$, whereas the entry
$\NoisyReward_k(\state, \action)$ is a zero-mean random variable with
mean $\reward(\state, \action)$ and $\sigma_r$-sub-Gaussian tails,
corresponding to a noisy observation of the reward function.  Here the
rewards $\{\NoisyReward_k(\state, \action)\}_{(\state, \action) \in
  \stateset \times \actionset}$ are independent across the all
state-action pairs, and the random rewards $\{\NoisyReward_k\}$ are
independent of the randomness in $\{\obsmat_k\}$.

Based on the observations, our goal is to estimate the optimal
$Q$-value function $\Qstar$, along with an optimal policy $\pistar$.
From the classical theory of MDPs~\citep{puterman2014markov,
  sutton1998, Bertsekas2009}, the optimal $Q$-function is a fixed
point of the Bellman (optimality) operator $\Op$, a map from
$\RR^{\card{\stateset} \times \card{\actionset}}$ to itself given by
\begin{align}
\label{EqnPopulationBellman}
\Op(\qvalues)(\state, \action) & \defn \reward(\state, \action) +
\contractPar \sum_{\state' \in \stateset} \Prob_\action(\state' \mid
\state) \max_{\action' \in \actionset} \qvalues(\state', \action'),
\end{align}
and an optimal policy $\policystar$ can be obtained from the optimal
$Q$-function $\Qstar$ via the maximization $\policystar(\state) \in
\arg \max \limits_{\action \in \actionset} \qvaluesstar(\state,
\action)$.  In this paper, we measure the quality of a given estimate
$\widehat{\qvalues}$ in terms of the $\ell_\infty$-norm error:
\begin{align}
  \inftynorm{\widehat{\qvalues} -
  \qvaluesstar} = \max_{(\state, \action)} |\widehat{\qvalues}(\state,
  \action) - \qvaluesstar(\state, \action)|.
\end{align}

\subsection{Contributions of this paper}

The main contribution of this paper is to show that for a given MDP, the
difficulty of estimating the optimal $Q$-value function in
$\ell_\infty$-norm is characterized by a particular functional of the
problem instance $(\TranMatQ, \reward)$, defined here.

\paragraph{An instance-dependent functional:}  Given a
sample $(\obsmat, \NoisyReward)$ from our observation model,
we can define the single-sample empirical Bellman operator
\begin{align}
  \label{EqnEmpiricalBellman}
  \NoisyOp(\qvalues) & \defn \NoisyReward(\state, \action) +
  \contractPar \sum_{\state' \in \stateset} \obsmat_\action(\state'
  \mid \state) \max_{\action' \in \actionset} \qvalues(\state',
  \action'),
\end{align}
where we have introduced $\obsmat_\action(\state' \mid \state) \defn
\Indicator_{ \obsmat(\state, \action) = \state'}$.

Note that for any fixed $Q$-function $\qvalues$, the difference
$\NoisyOp(\qvalues) - \Op(\qvalues)$ is a zero-mean random matrix, and
a key object in this paper is the matrix $\AllCompTerm \in
\RR^{\card{\stateset} \times \card{\actionset}}$ with entries
\begin{align}
  \label{EqnKey}
\AllCompTerm(\policy; \instance, \contractPar)(\state, \action) &
\defn \sqrt{\var \Big( \big(\Id - \contractPar \TranMatQ^\pi
  \big)^{-1} \big( \NoisyOp(\Qstar) - \Op(\Qstar) \big) \Big)}.
\end{align}
More explicitly, the quantity $\TranMatQ^{\pi}$ is a
right-linear mapping of $\RR^{\card{\stateset} \times
  \card{\actionset}}$ to itself, given by:
\begin{align}
\label{EqnTransMat}
\TranMatQ^\policy \Q(\state, \action) & \defn \sum_{\state' \in
  \stateset} \TranMatQ_{\action}(\state' \mid \state) \cdot
\Q(\state', \policy(\state')) \qquad \mbox{for each $(\state, \action)
  \in \stateset \times \actionset$,}
\end{align} 
and the square-root and variance operators in equation~\eqref{EqnKey}
are applied elementwise.

Let us provide some intuition as to why $\AllCompTerm(\policy;
\instance, \contractPar)$ plays a fundamental role.  The appearance of
the zero mean term $\NoisyOp(\Qstar) - \Op(\Qstar)$ is natural: it
reflects the noise present in the empirical Bellman
operator~\eqref{EqnEmpiricalBellman} as an estimate of the population
Bellman operator~\eqref{EqnPopulationBellman}.  As for the pre-factor
$(\Id - \contractPar \TranMatQ^{\pi})^{-1}$, by a von Neumann
expansion we can write
\begin{align*}
(\Id - \contractPar \TranMatQ^{\pi})^{-1} & = \sum_{k=0}^\infty
(\contractPar \TranMatQ^\pi)^k.
\end{align*}
The sum of the powers of $\contractPar \TranMatQ^\pi$ account for the
compounded effect of an initial perturbation when following the Markov
chain specified by the policy $\pi$.

\paragraph{Upper and lower bounds:}

With these definitions in place, the core of our work involves
proving, via a combination of a lower and an upper bound (matching up
to logarithmic factors), that the instance-specific difficulty of
estimating the $Q$-function is captured by the quantity $\max_{\pi \in
  \OptPolicySet} \inftynorm{\AllCompTerm(\pi; \instance,
  \contractPar)}$.  Here $\OptPolicySet$ denotes the set of all
optimal policies for the MDP instance $(\instance)$.  This functional
exhibits a wide range of behaviors: in Example~\ref{ExaSimple} to
follow in Section~\ref{SecExplore}, we exhibit a family of MDPs
$(\TransMatQ_\lampar, \reward_\lampar)$, parameterized by a scalar
$\lampar \geq 0$ such that
\begin{align*}
\max_{\pi \in \OptPolicySet} \inftynorm{\AllCompTerm(\pi;
  \TransMatQ_\lampar, \reward_\lampar, \contractPar)} & \asymp
\Big(\frac{1}{1- \contractPar} \Big)^{\tfrac{1}{2} - \lampar}.
\end{align*}
The setting $\lampar = 0$ recovers a ``hard'' instance, one for which
the global minimax bound for estimation of $Q$-functions, known from
past work~\cite{Azar2013Minimax} on batched $Q$-learning, is sharp.
On the other hand, as $\lampar$ grows, the problems in this family
become progressively easier, so that the global minimax bound is no
longer sharp.

In more detail, we prove a non-asymptotic lower bound, stated as
Theorem~\ref{ThmLower} to follow, by adapting a particular definition
of local minimax risk studied in past work on shape-constrained
estimation~\cite{cailow2015estconv}.  The central challenge in this
proof is that perturbations to the transition matrices of a given MDP
change not only the transitions themselves, but also the structure of
the optimal policies.  In order to prove matching upper bounds, given
the role of the empirical operators $\NoisyOp$ in our lower bound,
which are used in the classical $Q$-learning
algorithm~\citep{Watkins1992Qlearning,tsitsiklis1994,
  szepesvari1997asymptotic, Jaakkola1994@StochDP}, a natural thought
would be to analyze this operator directly.  However, it is known from past
work~\cite{wainwright2019stochastic} that the classical $Q$-learning
algorithm is \emph{non-optimal}, even when assessed when using the
coarser metric of global minimax.  Thus, in order to obtain a sharp
upper bound, we turn to variance-reduced forms of $Q$-learning, as
introduced in past
work~\citep{sidford2018near,sidford2018variance,wainwright2019variancereduced}
and shown to be optimal in a globally minimax sense.  Our main
contribution is to show that under certain structural conditions and lower
bounds on the sample size, there is a form of variance-reduced
$Q$-learning that achieves our local minimax lower bound up to a
logarithmic factor.  These upper bounds, stated precisely in
Theorem~\ref{ThmUpper},  confirm that our lower bound technique has
extracted a useful form of instance dependence for estimating optimal
$Q$-functions.

\paragraph{Notation:} For a positive integer $n$, we use the
shorthand $[n] \defn \{1, 2, \ldots n\}$. For a finite set $S$, we use
$\card{S}$ to denote its cardinality. We use $c_1, c_2, \ldots$ to
denote universal constants that may change from line to line. For any
pair of vectors or matrices $(v, w)$ with matching dimension(s), we
write that $v \succeq w$ to imply $v - w$ has only positive entries,
and $v \preceq w$ is defined similarly. We let $|u|$ denote the
entrywise absolute value of a vector $u \in \RR^n$ or a matrix $u \in
\RR^{m \times n}$; we use $\abs{u}_{+}$ to denote the entry-wise
positive part of $u$. For any vector or matrix $u$, we let
$\inftynorm{u}$ denote the maximum absolute value taken over all
entries of $u$, and $\VecSpan{u} = \max_j u_j - \min_j u_j$ denote the
span seminorm. For a continuous operator $P: \RR^{m \times n} \to
\RR^{m \times n}$, we define its $\ell_\infty$-operator norm as
$\matnorminfty{P} = \sup_{\|u\|_\infty = 1} \inftynorm{Pu}$.  We often
identify a $Q$-value function $\Q$ with its matrix representation and
use $\inftynorm{\Q}$ to denote the infinity norm (i.e., largest entry
in absolute terms).  In the matrix representation of $\Q$, its rows
and columns are indexed via an enumeration of the states and actions,
respectively. We use the symbol $\gtrsim$ to denote a relation that
holds up to logarithmic factors in the problem parameters.


\section{Main results}
\label{SecMain}

We proceed to provide precise statements of the main results of
this paper, along with a discussion of some of their consequences.  In
Section~\ref{SecLower}, we define a notion of a local non-asymptotic
minimax risk, and then state Theorem~\ref{ThmLower}, which provides
such a lower bound for estimating optimal $Q$-value functions.  In
Section~\ref{SecUpper}, we turn to the complementary problem of
deriving achievable results.  Theorem~\ref{ThmUpper} shows that under
certain structural conditions on the policies, there is a form of
variance-reduced $Q$-learning that achieves the local minimax risk up
to logarithmic factors.


\subsection{Instance-dependent lower bounds}
\label{SecLower}

In this section, we state a non-asymptotic lower bound for estimating
optimal $Q$-function in the $\ell_\infty$-norm. This lower bound, to
be stated in Theorem~\ref{ThmLower}, is instance-dependent, meaning
that it depends on the particular instance of the MDP $(\TranMatQ,
\reward)$ at hand.  This dependence should be contrasted with
classical global minimax bounds, which are oblivious to such
local properties.

The starting point of our lower bound development is the two-point
framework introduced by Cai and Low~\citep{cailow2015estconv} for
local minimax bounds for nonparametric shape-constrained inference;
here we adapt it to our current setting. Focusing on the
$\ell_\infty$-norm error metric, the \textit{local non-asymptotic
  minimax risk} for $\qvalues(\probinstance)$ at an instance
$\probinstance =(\Trans, \reward)$ is defined as
\begin{align}
\label{eqn:lower-bound-def}
\minimax_\Qnumobs(\probinstance) \defn \sup_{\probinstance'}
\inf_{\qestimate} \max_{\mathcal{Q} \in \{\probinstance,
  \probinstance'\}} \sqrt{\Qnumobs} \cdot
\Exs_{\mathcal{Q}}\left[\inftynorm{\qestimate - \qvalues(\mathcal{Q})}
  \right].
\end{align}
Here the infimum is taken over all estimators $\qestimate$ that are
measurable functions of the $\Qnumobs$ i.i.d.\ observations drawn
according to our observation model (see Section~\ref{SecSomeBack}).

The intuition underlying the definition~\eqref{eqn:lower-bound-def} is
that given an instance $\probinstance$, the adversary that defines the
instance-dependent non-asymptotic risk
$\minimax_\Qnumobs(\probinstance)$ behaves as follows: it extracts
the hardest alternative $\probinstance'$ relative to $\probinstance$,
and then measures the worst-case risk over $\probinstance$ and this
alternative $\probinstance'$.

\subsubsection{Lower bounds for $Q$-function estimation}

We now turn to the statement of some lower bounds for estimating the
optimal $Q$-function.  Recall the definition~\eqref{EqnTransMat} of
the operator $\TransMatQ^\pi$, along with the functional
$\AllCompTerm(\policy; \Trans, \reward, \discount)$ from
equation~\eqref{EqnKey}.  We let $\AllCompTerm^2(\policy; \Trans,
\reward, \discount)$ denote the matrix obtained by taking squares
entrywise.  Our first step is to provide a decomposition of this
matrix into two separate components, corresponding to the noisiness in
the reward function observation and transition matrix observations,
respectively.

In order to deal with the latter source of noise, with a slight abuse
of notation, we use the observed matrix $\obsmat$ to define a
stochastic analog of $\TransMatQ^\pi$---namely, the (random)
right-linear operator
\begin{align}
\label{eqn:Sample-Tran-PI}
(\obsmat^\policy \qvalues)(\state, \action) & \defn \sum_{\state' \in
  \stateset} \obsmat_\action(\state' \mid \state) \cdot
\qvalues(\state', \policy(\state')), \quad \text{where} \quad
\obsmat_\action(\state' \mid \state) \defn \Indicator_{
  \obsmat(\state, \action) = \state'}.
\end{align}

By assumption, the randomness in our observations of the reward and
transitions are independent, so that for any
optimal\footnote{Optimality of $\pi$ is required so that
$\Op(\Qstar) = \reward + \contractPar \TransMatQ^\pi \Qstar$,
with a similar relation for the empirical Bellman operator.}  policy
$\pi$, we have the decomposition
\begin{subequations}
\begin{align}
\label{eqn:variance-decomposition}
\AllCompTerm^2(\policy; \Trans, \reward)(\state, \action) =
\contractPar^2 \CompTerm^2(\policy; \Trans, \reward)(\state, \action)
+ \RewardTerm^2(\policy; \Trans, \reward)(\state, \action).
\end{align}
Here we define
  \begin{align}
  \label{eqn:CompTerm}
  \CompTerm^2(\policy; \Trans, \reward) & \defn \var\left((\Id -
  \contractPar \Trans^\policy)^{-1} (\obsmat^{\policy} -
  \TranMatQ^{\policy}) \Qstar \right), \qquad \text{and} \\
  \label{eqn:RewardTerm}  
  \RewardTerm^2(\policy; \Trans, \reward) & \defn \var\left((\Id -
  \contractPar \Trans^\policy)^{-1} (\NoisyReward - \reward)\right),
  \end{align}
\end{subequations}
where we compute the variances in an elementwise sense. \\

With this notation, we have the following guarantee:
\begin{theorem}
\label{ThmLower}
There exists a universal constant $c > 0$ such that for any instance
$\probinstance = (\Trans, \reward)$, the local non-asymptotic minimax
risk is lower bounded as
\begin{subequations}
\begin{align}
\label{eqn:Main-LB}
\minimax_\Qnumobs(\probinstance) \geq c \max_{\policy \in
  \OptpolicySet} \inftynorm{\AllCompTerm(\policy; \instance,
  \contractPar)}.
\end{align}
This bound is valid for all sample sizes $\Qnumobs$ that satisfy the
lower bound
\begin{align}
\label{eqn:min-sample-size-requirement}
\Qnumobs \geq \Qnumobs_0 \defn \max \left\{\frac{2\contractPar^2}{(1 -
  \contractPar)^2}, \, \frac{2 \VecSpan{\qvaluesstar}^2}{(1 -
  \contractPar)^2 \inftynorm{\CompTerm^2(\pistar; \Trans, \reward)}}
\right\},
\end{align}
\end{subequations}
where $\pistar \in \arg \max_{\policy \in \OptpolicySet}
\inftynorm{\AllCompTerm(\policy; \instance)}$.
\end{theorem}
\noindent We prove this theorem in Section~\ref{SecProofLower}.  The
main take-away is that the functional $\max \limits_{\policy \in
  \OptpolicySet} \inftynorm{\AllCompTerm(\policy; \instance,
  \contractPar)}$ controls the local minimax risk.  In order to gain
intuition for this claim, it is worth exploring the range of possible
behaviors exhibited by this functional.


\subsubsection{Exploring the range of possible behaviors}
\label{SecExplore}

One point of comparison is between the instance-dependent lower bound
from Theorem~\ref{ThmUpper} with the existing minimax lower bounds for
$Q$-learning.  Azar et al.~\citep{Azar2013Minimax} provided a global
minimax lower bound on the $\ell_\infty$-norm error for estimating the
optimal $Q$-function.  For a $\contractPar$-discounted MDP, they
showed that the $\ell_\infty$-error of any procedure is lower bounded
by the quantity $\frac{1}{(1 - \contractPar)^{1.5}} \cdot
\frac{1}{\sqrt{\numobs}}$, up to logarithmic factors in dimension.

This lower bound is optimal in a globally minimax sense, and it is
worthwhile understanding the properties of instances that exhibit this
worst-case behavior---that is, instances for which $\max_{\policy \in
  \OptpolicySet} \inftynorm{\AllCompTerm(\policy; \instance,
  \contractPar)} \asymp \tfrac{1}{(1-\contractPar)^{1.5}}$.  It is
also worthwhile understanding the properties of problems that are much
``easier'' than this worst-case theory would suggest.  The following
construction, which takes inspiration from~\citep{pananjady2020instancedependent,khamaru2020PE}, allows us
to explore this continuum.

\begin{example}[A continuum of local minimax risks]
  \label{ExaSimple}
Consider an MDP with two states $\{\state_1, \state_2\}$, two actions
$\{\action_1, \action_2\}$, and with transition functions and reward
functions given by
\begin{align}
\label{eqn:ex-1-transitions-andrewards}
    \probmat_{\action_1} = \begin{bmatrix} p & 1 - p \\ 0 &
      1 \end{bmatrix} \quad \probmat_{\action_2} = \begin{bmatrix} 1 &
      0 \\ 0 & 1 \end{bmatrix}, \quad \mbox{and} \quad \reward
    = \begin{bmatrix} 1 & 0 \\ \tau & 0 \end{bmatrix}.
\end{align}
We assume that there is no randomness in the rewards. Here, the pair
$(p, \taupar)$ along with the discount factor $\contractPar$ are
parameters of the construction, and we consider a sub-family of these
parameters indexed by a scalar $\lampar \geq 0$.  For any such
$\lampar$ and discount factor $\contractPar \in (\tfrac{1}{4}, 1)$,
consider the settings
\begin{align*}
    p = \frac{4 \contractPar - 1}{3 \contractPar}, \quad \text{and}
    \quad \taupar = 1 - (1 - \contractPar)^{\lampar}.
\end{align*}
With these choices, the optimal $Q$-function $\Qstar$ takes the form
\begin{align*}
\Qstar = \begin{bmatrix} \frac{1}{4} \cdot \frac{3 + \taupar}{1 -
    \contractPar} & \frac{\contractPar}{4} \cdot \frac{3 + \taupar}{1
    - \contractPar} \\ \frac{\taupar}{1 - \contractPar} &
  \frac{\contractPar \taupar}{1 - \contractPar} \end{bmatrix},
\end{align*} 
with an unique optimal policy $\policystar(\state_1) =
\policystar(\state_2) = \action_1$.  We can then compute that
\begin{align}
\label{eqn:simple-instance-bound}
    \max_{\pistar \in \OptpolicySet} \inftynorm{ \Nu(\pistar ;
      \lambdaInstance)} = c \cdot
    \left(\frac{1}{1-\contractPar}\right)^{1.5 - \lampar}.
\end{align}
See Appendix~\ref{sec:simple-problem-details} for the details of this
calculation.

Substituting into equation~\eqref{eqn:Main-LB} yields
that the local minimax risk is lower bounded as
\mbox{$\minimax_\Qnumobs(\probinstance) \geq c\frac{1}{(1 -
    \contractPar)^{1.5 - \lampar}}$.}  Consequently, for $\lampar >
0$, our lower bounds suggest it should be possible to estimate the
optimal $Q$-function more accurately by a factor $(1 -
\contractPar)^\lampar$; note that this difference is particularly
significant for values of the discount factor $\contractPar$ that are
close to one. 
\end{example}


\subsection{Instance-dependent upper bounds}
\label{SecUpper}

Thus far, we have stated some instance-dependent lower bounds on the
sample complexity of estimating $Q$-value functions.  As we saw in the
preceding Example~\ref{ExaSimple}, these lower bounds exhibit a wide
range of behavior depending on the structure of the transition
functions, discount parameter and reward functions.  However, these
differences in the lower bounds are only interesting if we can show
that they are optimal, meaning that there is a (hopefully practical)
algorithm that matches the behavior predicted by the lower bounds.

In this section we close this gap, in particular via a careful
analysis of variance-reduced $Q$-learning (or~\ref{AlgVR} for short).
Variance-reduced forms of $Q$-learning have been proposed and shown to
be globally minimax in previous
work~\citep{sidford2018near,sidford2018variance,wainwright2019variancereduced};
the version analyzed here is motivated from~\citep{wainwright2019variancereduced}.  In
Theorem~\ref{ThmUpper}, we show that the~\ref{AlgVR} algorithm is
instance-optimal up to logarithmic factors under two different sets of
assumption.


\subsubsection{From standard to variance-reduced $Q$-learning}

The classical $Q$-learning algorithm is a stochastic approximation
algorithm for estimating the unique fixed point $\Qstar$ of the
Bellman operator $\Op$.  Recall the
definition~\eqref{EqnEmpiricalBellman} of the empirical Bellman
operator $\NoisyOp_k$. At each iteration $k = 1, 2, \ldots$, standard
$Q$-learning performs an update of the form
\begin{align}
  \label{EqnStandardQ}
  \Q_{k + 1} = (1 - \stepsize_k) \Q_k + \stepsize_k \NoisyOp_k(\Q_k),
\end{align}
where $\stepsize_k \in (0, 1)$ is a stepsize parameter.  Appropriately
decaying choices of the stepsize ensure that the estimate $\Q_k$
converges to $\Qstar$.  Unfortunately, the convergence rate is known
to be non-optimal, failing to achieve the global minimax
rate~\cite{wainwright2019stochastic}, let alone the finer-grained
instance-dependent requirements in this paper.  This non-optimality
has to do with the rate at which variance accumulates as the procedure
is run.

Variance reduction is a general principle that can be applied to
stochastic approximation schemes so as to accelerate their
convergence.  Here we describe the variance-reduced version of
$Q$-learning that we analyze here.  Similar to standard
variance-reduced schemes for stochastic optimization~\citep[see, e.g.,][]{johnson2013accelerating}, the algorithm consists of a
sequence of epochs.  Within each epoch, we run a re-centered version
of the \QL update. The re-centering is done in such a way, using a
Monte Carlo approximation of the population Bellman operator $\Op$, so
that the re-centered updates have lower variance.  We leave the details
of the epochs and Monte Carlo to Section~\ref{SecDetails}; here let us
describe the basic form of the updates within a given epoch.

Suppose that we run the algorithm using a total of $\epochs$ epochs.
At epoch $m$, the algorithm uses a re-centering point $ \elebar_m$ in
order to re-center the update, where $\elebar_m$ acts as the current
best estimate of $\elestar$. Ideally, we should re-center the operator
$\NoisyOp_k$ using the quantity $\Op(\elebar_m)$, but we lack the
access to it; instead, we use the Monte Carlo approximation
\begin{align}\label{EqnMCRecenter}
\RecenterOp(\elebar_m) \defn \frac{1}{\recentersize} \sum_{i \in \recentersample} \NoisyOp_i(\elebar_m).
\end{align}
Given the pair $(\elebar_m, \RecenterOp(\elebar))$ and a stepsize
parameter $\stepsize \in (0,1)$, we define the \emph{variance-reduced
$Q$-learning update} as follows:
\begin{align}
 \label{EqnVROpDef}
\begin{split}
\ele & \mapsto \VROp_k \left(\ele; \stepsize, \elebar_m,
\RecenterOp\right) \defn (1 - \stepsize) \ele + \stepsize
\left\{\NoisyOp_k(\ele ) - \NoisyOp_k(\elebar_m) +
\RecenterOp(\elebar_m)\right\},
\end{split}
\end{align}
where the operator $\NoisyOp_k$ is independent of the set of operators
$\{\NoisyOp_i\}_{i \in \recentersample}$, used to compute the Monte
Carlo approximation $\RecenterOp$. As a result, the stochastic
operator $\NoisyOp_k$ is independent of the re-centering quantity
$\RecenterOp(\elebar_m)$.  See Section~\ref{SecDetails} for the
details on how the epoch lengths and re-centering sample sizes
$\recentersample$ are chosen.

\subsubsection{Non-asymptotic guarantees for variance-reduced $Q$-learning}

In this section, we state some non-asymptotic guarantees for
the \VRQLspace algorithm.  We provide guarantees under two conditions,
both of which involve the structure of the set of optimal policies.
We begin by introducing some definitions that underlie these two
conditions.

Given an MDP instance $(\TranMatQ, \reward)$, we define the
\emph{optimality gap}
\begin{align}
  \label{eqn:sub-optimality}
    \SpGap & \defn \min_{\policy \in \PolicySet \setminus
      \OptpolicySet} \inftynorm{\Qstar - \{ \reward +
      \contractPar\TranMatQ^{\policy}\Qstar \}},
\end{align}
where $\Qstar, \OptPolicySet$, and $\PolicySet$, respectively, denote
the optimal $Q$-function, the set of optimal policies, and the set of
all policies for MDP $(\instance)$. Observe that the scalar $\SpGap$
captures the difficulty in detecting the set of optimal policies. In
other words, when $\SpGap$ is small, it is hard to distinguish an
optimal policy from a suboptimal policy.

Our second set of conditions involves the family of right-linear
operators $\{\TransMatQ^\pi \;:\; \pi \in \PolicySet \}$ defined in
equation~\eqref{EqnTransMat}.  For any $Q$-value function $\ele$, we
say that a policy $\pi$ is greedy with respect to $\ele$ if
$\pi(\state) \in \arg \max_{\action \in \ActionSpace} \ele(\state,
\action)$ for all $\state \in \StateSpace$.  Note that any policy
$\pistar$ that is greedy with respect to the optimal $Q$-value
function $\Qstar$ is an optimal policy.  We say that the operators
satisfy a \emph{Lipschitz condition} if there is a constant $L$
such that for any $Q$-value function $\ele$ and associated
greedy-optimal policy $\pi$, we have
\begin{align}
\label{EqnLipschitz}
\inftynorm{(\TranMatQ^{\policy} - \TranMatQ^{\policystar}) \; (\ele -
  \Qstar)} \leq L \inftynorm{\ele - \Qstar}^2.
\end{align}
Intuitively, this condition means that the operator difference
$\TransMatQ^\policy - \TransMatQ^{\policystar}$ is small whenever the
underlying $Q$-value functions that induce the policies are close.
Conditions of this type were introduced by Puterman and
Brumelle~\cite{putermanbrumelle1979} in their classical analysis of the convergence rates
of policy iteration algorithms for exact dynamic programming.

With these definitions in place, we can specify the two settings under
which we provide upper bounds on the \VRQLspace algorithm:

\begin{description}
\item[Setting \UNQ:] There is a unique optimal policy $\pistar$, and
  the sample size $\Qnumobs$ is lower bounded as
\begin{align}
   \label{eqn:N-lb-for-instance-opt}
   \frac{\Qnumobs}{(\log\Qnumobs)^2} \geq c_2 \log(\numstates /
   \pardelta) \cdot \frac{(1 + \inftynorm{\reward} + \sigma_\reward
     \sqrt{1 - \contractPar})^2}{(1 - \contractPar)^3} \cdot \max\{1, \,
   \frac{1}{\SpGap^2 ( 1 - \contractPar)^\nLBBeta} \} \quad \text{for
     some} \;\; \nLBBeta > 0.
 \end{align}    
\item[Setting \LIP:] The Lipschitz condition~\eqref{EqnLipschitz}
  condition holds, and the sample size is lower bounded as
\begin{align}
\frac{\Qnumobs}{(\log \Qnumobs)^2} \geq c_2 \log(\numstates/\pardelta)
\frac{( 1 + \inftynorm{\reward} + \sigma_\reward\sqrt{1 -
    \contractPar})^2}{(1 - \contractPar)^{3 + \beta}} \cdot \min\{\frac{L^2}{(1 - \contractPar)^2}, \, \frac{1}{\SpGap^2} \} \qquad \mbox{for
  some $\beta > 0$.}
\end{align}
\end{description}

In all cases, we assume that we are given an initial point $\Qbar_1$
such that
\begin{align}
  \label{eqn:Init-cond-smallsample}
  \infnorm{\Qbar_1 - \Qstar} \leq \frac{\infnorm{r}}{\sqrt{1 -
      \contractPar}}.
\end{align}
Such an initial condition has already been used in the literature~\cite{wainwright2019variancereduced}, and it can be ensured by first running Algorithm~\ref{AlgVR} for a total of $\frac{1}{(1 - \contractPar)^3}$ samples (up to logarithmic factor corrections). 

\begin{theorem}
  \label{ThmUpper}
Under either settings (UNQ) or (LIP), there are choices of epoch
parameters such that given any discount parameter $\contractPar \in
[\frac{1}{2}, 1)$ and an initial point $\Qbar_1$ satisfying the
  initialization condition~\eqref{eqn:Init-cond-smallsample},
  Algorithm~\ref{AlgVR} run for $\epochs \defn \log_4
  \left(\frac{\Qnumobs(1-\contractPar)^2}{8\log((16\numstates/\pardelta)
    \cdot \log \Qnumobs)}\right)$ epochs yields an estimate
  $\Qbar_{\epochs + 1}$ such that
\begin{align}
\label{eqn:matching-upper-bound}
    \inftynorm{\Qbar_{\epochs + 1} - \Qstar} &\leq c_0 \cdot
    \sqrt{\frac{\LogDM}{\Qnumobs}} \CompTermInstance + c_1 \cdot
    \frac{\LogDM}{\Qnumobs} \cdot \frac{\VecSpan{\Qstar}}{1 -
      \contractPar},
\end{align}
with probability exceeding $1-\delta$.
\end{theorem}
\noindent See Section~\ref{SecProofUpper} for the proof of this claim.

\paragraph{Comparing the upper and lower bounds:} Assuming the sample size lower bound~ from Theorem~\ref{ThmLower} are valid, we see that the second term in the bound~\eqref{eqn:matching-upper-bound} is of smaller order. In this case, the upper bound from Theorem~\ref{ThmUpper} and the lower bound from Theorem~\ref{ThmLower} matches, and we conclude that the~\VRQL~algorithm is \emph{instance optimal}.

Although the guarantee~\eqref{eqn:matching-upper-bound} involves the
same $1/\sqrt{\numobs}$ rate and complexity term $\CompTermInstance$ as the
lower bound in Theorem~\ref{ThmLower}, it should be noted that the
sample size lower bounds required for Theorem~\ref{ThmUpper} are
more stringent than that in Theorem~\ref{ThmLower}.  Moreover, our
lower bound does not require the side conditions---either the unique optima
or Lipschitz conditions---that are imposed in Theorem~\ref{ThmUpper}.
Closing these remaining differences between the two results is a worthwhile goal for future work.


\subsubsection{Confirming the theoretical predictions}

Some numerical experiments are helpful in order to illustrate
instance-adaptive behavior guaranteed by Theorem~\ref{ThmUpper}.
Recall the family of MDPs~\eqref{eqn:ex-1-transitions-andrewards} from
Example~\ref{ExaSimple}. Suppose that we set $\lampar = 0.5$ and for
each choice of $\contractPar \in (1/2, 1)$, we collect $\Qnumobs =
\lceil \frac{16 \cdot 32}{9} \frac{1}{(1 - \contractPar)^3 \cdot}
\rceil$ samples, and then run the \VRQLspace algorithm over a range of
discount parameters $\contractPar$, using the settings from
Theorem~\ref{ThmUpper} and Section~\ref{SecDetails}, thereby obtaining
an estimate $\Qbar_{\epochs + 1}$.

\begin{figure}[ht!]
  \begin{center}
    \begin{tabular}{ccc}
    \widgraph{0.4 \textwidth}{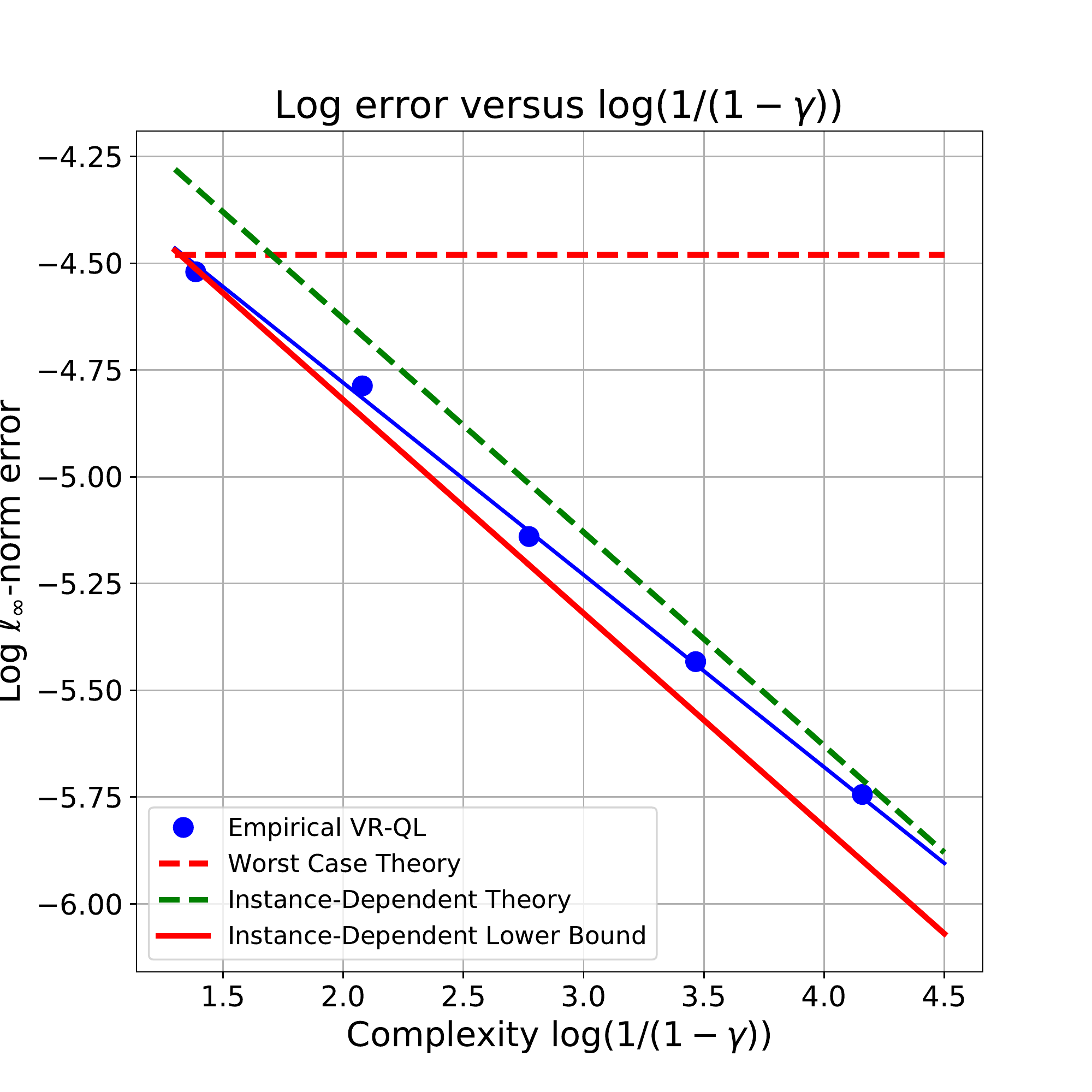} &&
    \widgraph{0.4\textwidth}{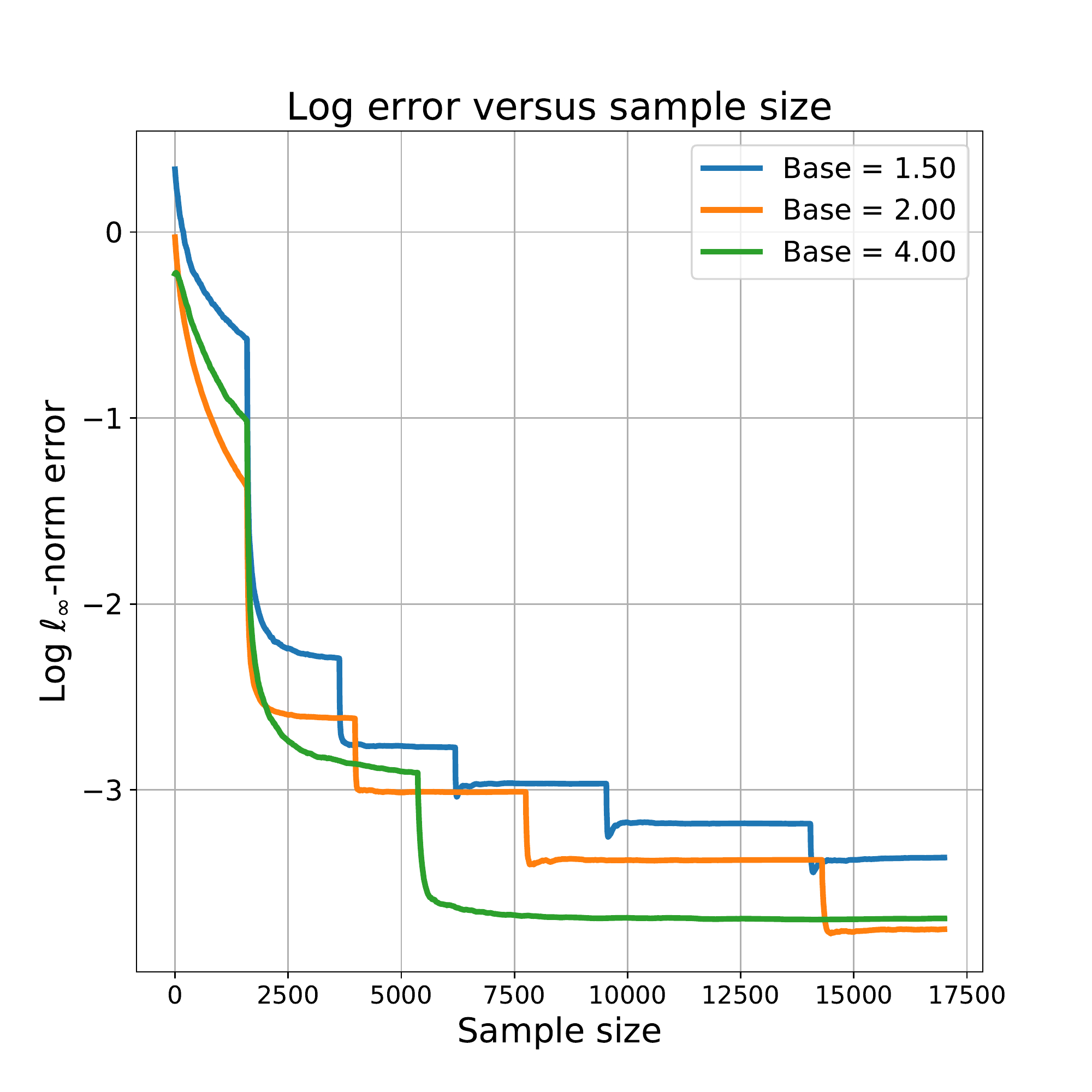} \\
    (a) && (b)
    \end{tabular}
    \caption{(a) $\lampar = 0.5, \quad \Qnumobs = \lceil \frac{32}{(1
        - \contractPar)^3 \cdot} \cdot \frac{4^2}{3^2} \rceil$
      $\contractPar = 0.9 $. Illustration of the qualitative behavior
      of Algorithm~\ref{AlgVR} applied on the MDP~\eqref{ExaSimple}
      along with instance dependent and the worst case bounds. The
      figure plots the $\log$ $\ell_\infty$-error
      $\inftynorm{\bar{\Q}_{\NumEpoch + 1} - \Qstar}$ against the log
      discount complexity factor $\log(\frac{1}{1 - \contractPar})$
      with $\lampar = 0.5$.  We have also plotted the least-squares
      fit through these points, and the instance-dependent lower bound
      from Theorem~\ref{ThmLower}, the instance-dependent upper bound
      from Theorem~\ref{ThmUpper}, and the worst-case
      bound~\citep{wainwright2019variancereduced}.  (b) Behavior of
      the~\VRQL~ algorithm with different choices of the base $b$. The
      plot demonstrates that different choices of the base $b$ yield
      similar behavior. }
\label{FigPlots}    
  \end{center}
\end{figure}

Figure~\ref{FigPlots}(a) plots the evolution of log $\ell_\infty$-norm
error of the estimate over time as the algorithm proceeds; the form of
these curves show the epoch-based nature of the convergence.  See
Section~\ref{SecDetails} for more details on the parameters of the
epochs, including the base parameter illustrated here.  Plotted as
blue circles in panel (b) of Figure~\ref{FigPlots} are the logarithm
of the $\ell_\infty$-norm error of the final output; that is, $\log
\inftynorm{\Qbar_{\epochs + 1} - \Qstar}$, versus the logarithm of the
discount complexity, $\log(1/(1-\contractPar))$.  Each point in this
plot represents an average over 1000 trials.

In terms of theory, with the settings given above, existing worst-case
bounds~\cite{Azar2013Minimax,wainwright2019variancereduced} predict
that the log $\ell_\infty$-norm error remains constant as the log
discount complexity grows; accordingly, we have plotted a dotted red
line with slope zero to illustrate the worst-case guarantee.  On the
other hand, for the MDP
instance~\eqref{eqn:ex-1-transitions-andrewards} with $\lampar = 0.5$,
a simple calculation yields that for the
instance~\eqref{eqn:ex-1-transitions-andrewards} the suboptimality gap
$\SpGap$ satisfies ${\SpGap = 1 - \frac{(1 - \contractPar)^\lampar}{4}
  \geq \frac{3}{4}}$. In our experiment, we set the sample size to be
$\Qnumobs = \lceil \frac{32}{(1 - \contractPar)^3} \cdot
\frac{4^2}{3^2} \rceil \geq \frac{32}{(1 - \contractPar)^3 \cdot
  \SpGap^2}$; as a result, the bounds from Theorems~\ref{ThmLower}
and~\ref{ThmUpper} are valid.

With the setting $\lampar = 0.5$, our calculations
from Example~\ref{ExaSimple} yield
\begin{align*}
\CompTermInstance & \asymp \Big( \tfrac{1}{1-\contractPar}
\Big)^{-0.5}.
\end{align*}
Thus, with the choice of sample size $\Qnumobs$ given above, our
theory predicts that the log $\ell_\infty$-norm error should exhibit
the scaling
\begin{align*}
  \log \inftynorm{\Qbar_{\epochs + 1} - \Qstar} \asymp \log \left(
  \frac{1}{\sqrt{\Qnumobs}} \CompTermInstance \right) \; \asymp \; c -
  0.5 \log \big(\tfrac{1}{1-\contractPar} \big),
\end{align*}
where $c$ is a constant.  In Figure~\ref{FigPlots}(b), we plot the
lower bound from Theorem~\ref{ThmLower} as a solid red line, and
the upper bound from Theorem~\ref{ThmUpper} as a dashed green line.
(While these lines both have slope $-0.50$, the intercept term $c$
is different due to the additional logarithmic factors in dimension
present in the upper bound.)

In order to test how the empirical behavior conforms to the
theoretical prediction, we did an ordinary least-squares fit of the
log $\ell_\infty$-norm error versus the log discount complexity; this
fit yields a line with slope $\betahat = -0.45$, and is plotted in
solid blue.  This test shows good agreement between the theoretical
prediction and the practical behavior.


\subsubsection{Details of the epochs and procedure}
\label{SecDetails}

In this section, we provide the complete details of the algorithm
used in our version of variance-reduced $Q$-learning.

\paragraph{A single epoch:}

A single epoch of the overall variance-reduced~\QL algorithm involves
repeated applications of the basic variance-reduced update $\VROp_k$
from equation~\eqref{EqnVROpDef}.  The epochs are indexed with
integers \mbox{$m = 1, 2, \ldots, \epochs$}, where $\epochs$
corresponds to the total number of epochs to be run. Each epoch $m$
requires the following four inputs:
\begin{itemize}
\item an element $\elebar$, which is chosen to be the output of the
  previous epoch $m - 1$;
\item a positive integer $\epochlength$ denoting the number of steps
  within the given epoch;
\item a positive integer $\recentersize$ denoting the batch size used
  to calculate the Monte Carlo update~\eqref{EqnMCRecenter};
\item a set of fresh operators $\{\NoisyOp_i\}_{i \in
  \epochsampleset}$, with \mbox{$|\epochsampleset| = \recentersize +
  \epochlength$}. The set $\epochsampleset$ is partitioned into two
  subsets having sizes $\recentersize$ and $\epochlength$,
  respectively. The first subset, of size $\recentersize$, which we
  call $\recentersample$, is used to construct the Monte Carlo
  approximation~\eqref{EqnMCRecenter}. The second subset, of size
  $\epochlength$ is used to run the $\epochlength$ steps within the
  epoch.
\end{itemize}

We summarize a single epoch in pseudocode form in
Algorithm~\ref{AlgRunEpoch}.
\begin{varalgorithm}{SingleEpoch}
\caption{\qquad RunEpoch $(\elebar; \epochlength, \recentersize,
  \{\NoisyOp_i\}_{i\in\epochsampleset})$}
\label{AlgRunEpoch}
\begin{algorithmic}[1]
\STATE Given (a) Epoch length $\epochlength$, (b) Re-centering vector
$\elebar$, (c) Re-centering batch size $\recentersize$, \mbox{(d)
  Operators} $\{\NoisyOp_i\}_{i \in \epochsampleset}$ \STATE Compute
the re-centering quantity
\begin{align*}
\RecenterOp(\elebar) \defn \frac{1}{\recentersize} \sum \limits_{i \in
  \recentersample} \NoisyOp_i(\elebar)
\end{align*}
\STATE Initialize $\Q_1 =  \Qbar$
\FOR{$k = 1, 2, \ldots, \epochlength$}
\STATE Compute the variance-reduced update:
\begin{align*}
  \ele_{k+1} &= \VROp_k(\ele_k; \stepsize_k, \elebar, \RecenterOp) \quad \text{with stepsize } \, \stepsize_k = \frac{1}{1 + (1 - \contractPar)k}.
\end{align*}
\ENDFOR
\STATE \textbf{return} $\ele_{\epochlength + 1}$
\end{algorithmic}
\end{varalgorithm}

\paragraph{Overall algorithm:}
The overall algorithm, denoted by~\ref{AlgVR} for short, has five
inputs: (a) an initialization $\Qbar_1$, (b) an integer $\NumEpoch$,
denoting the number of epochs to be run, (c) an integer
$\epochlength$, denoting the length of each epoch, (d) a sequence of
batch sizes $\{\recentersize\}_{m=1}^\epochlength$, denoting the
number of operators used for re-centering in the $\NumEpoch$ epochs,
and (e) sample batches $\{\{\NoisyOp_i\}_{i \in
  \epochsampleset}\}_{m=1}^\NumEpoch$ to be used in the $\NumEpoch$
epochs.  Given these five inputs, the overall procedure can be
summarized as in Algorithm~\ref{AlgVR}.

\begin{varalgorithm}{\VRQL}
\caption{}
\label{AlgVR}
\begin{algorithmic}[1]
\STATE Given (a) Initialization $\Qbar_1$, (b) Number of epochs,
$\epochs$,  (c) Epoch length $\epochlength$,  (d)
Re-centering sample sizes $\{\recentersize\}_{m=1}^M$, (e) Sample
batches $\{\NoisyOp_i\}_{i \in \epochsampleset}$ for \mbox{$m = 1,
  \ldots, \epochs$}
\STATE Initialize at $\elebar_1$ \FOR{$m = 1, 2, \ldots, \epochs$}
\STATE $\elebar_{m+1} = \text{RunEpoch}(\elebar_m; \epochlength,
\recentersize, \{\NoisyOp_i\}_{i \in \epochsampleset})$ \ENDFOR \STATE
\textbf{return} $\elebar_{M+1}$ as final estimate
\end{algorithmic}
\end{varalgorithm}


\paragraph{Settings for Theorem~\ref{ThmUpper}:}
Given a tolerance probability $\pardelta \in (0, 1)$ and the number of
available i.i.d.\ samples $\Qnumobs$, we run Algorithm~\ref{AlgVR} with
a total of $\epochs \defn \log_4
\left(\frac{\Qnumobs(1-\contractPar)^2}{8\log((16\numstates/\pardelta)
  \cdot \log \Qnumobs)}\right)$ epochs, along with the following
parameter choices: \\
\begin{subequations}\label{EqnQParam-small-sample}
\texttt{Re-centering sizes:}
\begin{align}\label{EqnPERecenterSize-smallsample}
\begin{split}
\Nm = c_1 \frac{4^\epoch}{(1-\contractPar)^2} \cdot
\log_4(16\epochs\numstates/\pardelta)
\end{split}
\end{align}
\texttt{Sample batches:}
\begin{align}\label{eqn:samplebatches-smallsample} 
\begin{split}
 &\text{Partition the $\Qnumobs$ samples to obtain $\{\NoisyOp_i\}_{i \in \epochsampleset}$ for $m = 1, \ldots, \epochs$}
\end{split}
\end{align}
\texttt{Epoch length:}
\begin{align}\label{eqn:Epoch-length-smallsample}
\epochlength = \frac{\Qnumobs}{2\epochs}.
\end{align}
\end{subequations}
%


\section{Proof of Theorem~\ref{ThmLower}}
\label{SecProofLower}

Given an MDP instance $\probinstance = (\Trans, \reward)$, we start by
introducing the following two classes of alternative MDPs:
\begin{align}
\Alter_1 = \{ \probinstance' = (\Trans', \reward') \mid \reward' = \reward \}, \quad \text{and} \quad \Alter_2 = \{ \probinstance' = (\Trans', \reward') \mid \Trans' = \Trans\}.
\end{align}
We consider the restricted version of the local minimax risk at the instance $\probinstance'$ to the classes $\Alter_i$:
\begin{align}
\minimax_{\Qnumobs}(\probinstance; \Alter_i) \defn \sup_{\probinstance' \in \Alter_i} \inf_{\hat{\qvalues}_\Qnumobs} \max_{\AltProb \in \{\probinstance, \probinstance'\}} \Exs_{\probdistr}[\sqrt{\Qnumobs} \inftynorm{\hat{\qvalues}_N - \qvalues(\AltProb)}], \quad i = 1, 2.
\end{align}
The main part of the proof involves showing that there exists a universal constant $c > 0$ such that 
\begin{subequations}
\label{eqn:statement}
\begin{align}
\label{eqn:statement-1}
\minimax_\Qnumobs(\probinstance; \Alter_1) &\geq c \cdot \max_{\policy \in \OptpolicySet} \inftynorm{ \contractPar \CompTerm(\policy; \Trans, \reward)}, \quad \text{and} \\
\label{eqn:statement-2}
\minimax_\Qnumobs(\probinstance; \Alter_2) &\geq c \cdot\max_{\policy \in \OptpolicySet} \inftynorm{ \RewardTerm(\policy; \Trans, \reward) },
\end{align}
\end{subequations}
where $\OptpolicySet$ denotes the optimal policy set for $(\Trans, \reward)$. We can then conclude
\begin{align*}
\begin{split}
\minimax_\Qnumobs(\probinstance) &\geq \max \{ \minimax_\Qnumobs(\probinstance; \Alter_1), \minimax_\Qnumobs(\probinstance; \Alter_2) \} \\
&\geq \frac{1}{2} \left(\minimax_\Qnumobs(\probinstance; \Alter_1) + \minimax_\Qnumobs(\probinstance; \Alter_2) \right) \\ 
&\geq \frac{c}{2} \max_{\policy \in \OptpolicySet} \inftynorm{ \contractPar \CompTerm(\policy; \Trans, \reward)} + \frac{c}{2} 
\max_{\policy \in \OptpolicySet} \inftynorm{\RewardTerm(\policy; \Trans, \reward)} \\
&\geq \frac{c}{2} \max_{\policy \in \OptpolicySet} \inftynorm{\AllCompTerm(\policy; \Trans, \reward)}.
\end{split}
\end{align*} 
The last inequality above follows from the
decomposition~\eqref{eqn:variance-decomposition}. It remains to prove
the claims~\eqref{eqn:statement-1} and~\eqref{eqn:statement-2}.  More
precisely, the core of our proof involves proving the following two
lemmas:
\begin{lemma}
\label{lem:minimax-to-Hellinger}
For all $\Alter \in \{ \Alter_1, \Alter_2 \}$, we have that
\mbox{$\minimax_\Qnumobs(\probinstance; \Alter) \geq \frac{1}{8}
  \underline{\minimax}_\Qnumobs(\probinstance; \Alter)$} where we
define
\begin{align*}
\underline{\minimax}_\Qnumobs(\probinstance; \Alter) \defn
\sup_{\probinstance' \in \Alter} \left\{\sqrt{\Qnumobs} \cdot
\inftynorm{\qvalues(\probinstance) - \qvalues(\probinstance')} \mid
\dhel{\probinstance}{\probinstance'} \leq \frac{1}{2\sqrt{\Qnumobs}}
\right\}.
\end{align*}
\end{lemma}
\noindent This lemma follows as a fairly straightforward consequence
of the standard reduction from estimation to testing; see
Appendix~\ref{proof:lem:minimax-to-Hellinger} for the details.\\

Our next lemma requires more effort to prove, and leverages the
specific structure of the problem at hand:
\begin{lemma}
\label{lem:Q-function-LB}
Given any MDP instance $\probinstance = (\TranMat, \reward)$:
\begin{itemize}
  \item[(a)] There exists an instance $\probinstance_1 = (\TranMat',
    \reward) \in \Alter_1$ such that
    $\dhel{\probinstance}{\probinstance_1} \leq
    \frac{1}{2\sqrt{\Qnumobs}}$ and
  \begin{align*}
     \sqrt{\Qnumobs} \cdot \inftynorm{\qvalues(\probinstance) -
       \qvalues(\probinstance_1)} \geq c \cdot \max_{\policy \in
       \OptpolicySet} \inftynorm{ \contractPar \CompTerm(\policy;
       \Trans, \reward)}.
\end{align*}
\item[(b)] There exists an instance $\probinstance_2 = (\TranMat,
  \reward') \in \Alter_2$ such that
  $\dhel{\probinstance}{\probinstance_2} \leq
  \frac{1}{2\sqrt{\Qnumobs}}$ and
  \begin{align*}
     \sqrt{\Qnumobs} \cdot \inftynorm{\qvalues(\probinstance) -
       \qvalues(\probinstance_2)} \geq c \cdot \max_{\policy \in
       \OptpolicySet} \inftynorm{\RewardTerm(\policy; \Trans,
       \reward)} .
\end{align*}
\end{itemize}
\end{lemma}
Note that the bounds~\eqref{eqn:statement-1}--\eqref{eqn:statement-2}
stated in Theorem~\ref{ThmLower} follow by combining
the claims of Lemmas~\ref{lem:minimax-to-Hellinger}
and~\ref{lem:Q-function-LB}.  The remainder of our proof focuses on
establishing Lemma~\ref{lem:Q-function-LB}.


\subsection{Proof of Lemma~\ref{lem:Q-function-LB}}
\label{proof:Lemma-Q-function-LB}

In this section, we prove the two parts of
Lemma~\ref{lem:Q-function-LB}.

\subsubsection{Proof of Lemma~\ref{lem:Q-function-LB}(a)}
Throughout the proof, we use $\sapair$ to denote a generic element of
the state-action set $\stateset \times \actionset$. Let $\qvalues$ be
the true $Q$-function for the MDP $\probinstance = (\Pmat,
\reward)$. We adopt the shorthands
\begin{subequations}
\label{eqn:shorthands-transition-mat}
\begin{gather}
\OptPiOne \in \arg\max_{\policy \in \OptPolicySet} \inftynorm{\CompTerm(\policy; \Trans, \reward)}, \quad  \optsapair \in \arg \max_{\sapair \in \stateset \times \actionset}  \CompTerm(\OptPiOne; \Trans, \reward) ,  \quad
\CompTermEntry(\sapair) \defn  \CompTerm(\OptPiOne; \Trans, \reward)(\sapair) ,
 \\
\InvTerm \defn (\Id - \contractPar\Trans^\OptPiOne)^{-1} \quad \text{and} \quad 
\SigTerm^2(\sapair) \defn \text{Var}\left(\obsmat^\OptPiOne  \qvalues(\sapair)\right).
\end{gather}
\end{subequations}
To explain this notation, we chose $\OptPiOne$ to be the optimal policy that achieves the largest $\ell_\infty$-norm across $\CompTerm(\OptPi; \Trans, \reward)$ for optimal policies $\OptPi$, we let $\optsapair$ is the state-action pair index that achieves the maximal entry of $\CompTerm(\OptPiOne; \Trans, \reward)$, and we use $\CompTermEntry$ as convenient shorthand to refer to the values of $\CompTerm(\OptPiOne; \Trans, \reward)$. This choice of notation implies that 
\begin{align*}
\CompTermEntry(\optsapair) = \max_{\policy \in \OptPolicySet} \inftynorm{\CompTerm(\policy; \Trans, \reward)}.
\end{align*}
Additionally, note that $\InvTerm$ is a linear transformation from $\RR^{\card{\stateset} \times \card{\actionset}}$ to itself, so we can express the action of $\InvTerm$ on  $\qvalues$ as 
\begin{align*}
(\InvTerm \qvalues) (\sapair) = \sum_{\sapair' \in \stateset \times \actionset} \InvTermInd{\sapair}{\sapair'} \qvalues(\sapair').
\end{align*}
Note moreover that 
\begin{align}
\label{eqn:target-var-expression}   
\SigTerm^2(\sapair)
  = \sum_{\state'} \TransInd{\sapair}{\state'} (\qvalues(\state', \OptPiOne(\state')) - (\Trans^\OptPiOne \qvalues)(\sapair))^2
  \quad \text{and} \quad
  \CompTermEntry^2(\sapair) = \sum_{\sapair'} (\InvTermInd{\sapair}{\sapair'})^2 \SigTerm^2(\sapair').
\end{align}
With these definitions, we now define  $\PerTransInd{\sapair}{\statesec}$ as follows (we will prove that this choice is a valid probability transition kernel shortly):
\begin{align}
\label{eqn:Perturbed-Transition}
\PerTransInd{\sapair}{\statesec} = \TransInd{\sapair}{\statesec} + \frac{1}{\CompTermEntry(\optsapair) \sqrt{2\Qnumobs}} \TransInd{\sapair}{\statesec} \InvTermInd{\optsapair}{\sapair} \cdot (\qvalues(\statesec, \OptPiOne(\statesec)) - (\Trans^\OptPiOne \qvalues)(\sapair)).
\end{align}
Here, we have used the shorthand $\TransInd{\sapair}{\statesec} \equiv \Pmat_{\action}(y \mid \state)$, where $\sapair = (\state, \action) \in \stateset \times \actionset$.
Let $\qvalues \defn \qvalues(\Trans, \reward)$, and $\Perqvalues \defn \qvalues(\PerTrans, \reward)$ be the optimal $Q$ functions for MDP instances
$(\Trans, \reward)$ and $(\PerTrans, \reward)$ respectively. In the rest of the proof, we use the following properties of $\PmatBar$.
\begin{lemma}
  \label{lem:TransMat-properties}
For any MDP $\probinstance = (\Pmat, \reward)$ and the optimal policy $\OptPiOne$ defined in equation~\eqref{eqn:shorthands-transition-mat}, the corresponding $\PmatBar$ has the following properties:
\begin{itemize}
  \item[(a)] The $\PmatBar$ is a probability transition kernel.
  \item[(b)] The MDP instances $\probinstance = (\Pmat, \reward)$ and $\probinstance_1 = (\PmatBar, \reward)$ satisfy 
  $\dhel{\probinstance}{\probinstance_1} \leq \frac{1}{2\sqrt{\Qnumobs}}$
  and ${\opnorm{\PerTrans^\OptPiOne - \Trans^\OptPiOne} \leq \frac{1}{\sqrt{2\Qnumobs}}}$. 
  \item[(c)] Each entry of $(\Id - \contractPar \Trans^\OptPiOne)^{-1} 
  [\PerTrans^\OptPiOne - \Trans^\OptPiOne] \qvalues$ is non-negative. 
\end{itemize}
\end{lemma}
\noindent See Appendix~\ref{subsec:proof-TransMat-properties} for a
proof of this lemma.

\noindent 
Equipped with these tools, we are now ready to lower bound the norm
$\|\qvalues - \Perqvalues\|_\infty$.  The optimal $Q$-functions
$\qvalues$ and $\Perqvalues$ satisfy the following Bellman equations:
\begin{align}
\label{eqn:bellman-opt-cond}
\qvalues = \reward + \contractPar \Trans^\OptPiOne \qvalues \qquad
\text{and} \qquad \Perqvalues = \reward + \contractPar
\PerTrans^{\Perpolicy} \Perqvalues,
\end{align}
where $\OptPiOne \in \OptpolicySet$ is the optimal policy that achieves $\max_{\policy \in \OptpolicySet} \inftynorm{\CompTerm(\policy; \Trans, \reward)}$, and $\Perpolicy$ is an optimal policy for $(\bar{\Trans}, \reward)$. By the optimality of policy $\Perpolicy$ and the $Q$-function $\Perqvalues$, we have the entrywise inequality \mbox{$\PerTrans^{\Perpolicy} \Perqvalues \succeq \PerTrans^\OptPiOne \Perqvalues$}, which implies \mbox{$(\Id - \contractPar\PerTrans^\OptPiOne) \Perqvalues \succeq (\Id - \contractPar \PerTrans^{\Perpolicy}) \Perqvalues = \reward$}. Thus, using the identity \mbox{$A_1^{-1} - A_0^{-1} =  A_1^{-1}(A_0 - A_1) A_0^{-1}$} for invertible operators $A_0$ and $A_1$, we have
\begin{align*}
\begin{split}
\Perqvalues - \qvalues &\succeq \left[(\Id - \contractPar \PerTrans^\OptPiOne)^{-1} - (\Id - \contractPar \Trans^\OptPiOne)^{-1}\right] \reward \\
&= (\Id - \contractPar \PerTrans^\OptPiOne)^{-1} \left[(\Id - \contractPar \Trans^\OptPiOne) - (\Id - \contractPar \bar{\Trans}^\OptPiOne)\right] (\Id - \contractPar \Trans^\OptPiOne)^{-1} \reward \\
&= \contractPar (\Id - \contractPar \Trans^\OptPiOne)^{-1} \left[ \PerTrans^\OptPiOne - \Trans^\OptPiOne \right] (\Id - \contractPar \Trans^\OptPiOne)^{-1} \reward \\
&\quad + \contractPar \left( (\Id - \contractPar \PerTrans^\OptPiOne)^{-1} - (\Id - \contractPar \Trans^\OptPiOne)^{-1}\right)(\PerTrans^\OptPiOne - \Trans^\OptPiOne) (\Id - \contractPar \Trans^\OptPiOne)^{-1} \reward \\
&= \contractPar (\Id - \contractPar \Trans^\OptPiOne)^{-1} \left[ \PerTrans^\OptPiOne - \Trans^\OptPiOne \right] \qvalues \\
&\quad + \contractPar \left( (\Id - \contractPar \PerTrans^\OptPiOne)^{-1} - (\Id - \contractPar \Trans^\OptPiOne)^{-1}\right)(\PerTrans^\OptPiOne - \Trans^\OptPiOne) \qvalues,
\end{split}
\end{align*}
where the final equation follows from the Bellman optimality condition~\eqref{eqn:bellman-opt-cond}. Lemma~\ref{lem:TransMat-properties}(c) guarantees that the entries of \mbox{$(\Id - \contractPar \Trans^\OptPiOne)^{-1} [\PerTrans^\OptPiOne - \Trans^\OptPiOne] \qvalues$} are non-negative, 
and therefore we conclude
\begin{align}
\label{eqn:Q-value-Lower-bound-1}
\inftynorm{\bar{\qvalues} - \qvalues} \geq \contractPar \inftynorm{(\Id - \contractPar\Trans ^\OptPiOne)^{-1}[\bar{\Trans}^\OptPiOne - \Trans^\OptPiOne]\qvalues} - \contractPar \inftynorm{((\Id - \contractPar \bar{\Trans}^\OptPiOne)^{-1} - (\Id - \contractPar \Trans^\OptPiOne)^{-1})(\bar{\Trans}^\OptPiOne - \Trans^\OptPiOne) \qvalues}.
\end{align}
Consider the second term
$T_2 \defn \inftynorm{(\Id - \contractPar \PerTrans^\OptPiOne)^{-1} - (\Id -
  \contractPar \Trans^\OptPiOne)^{-1})(\PerTrans^\OptPiOne -
  \Trans^\OptPiOne) \qvalues}$.
We have
\begin{align*}
T_2 & = \inftynorm{[(\Id - \contractPar \PerTrans^\OptPiOne)^{-1}(\Id
    - \contractPar \Trans^\OptPiOne) - \Id] (\Id - \contractPar
  \Trans^\OptPiOne)^{-1}(\PerTrans^\OptPiOne -
  \Trans^\OptPiOne)\qvalues} \\
& \leq \opnorm{(\Id - \contractPar \PerTrans^\OptPiOne)^{-1}(\Id -
  \contractPar \Trans^\OptPiOne) - \Id} \cdot \inftynorm{(\Id -
  \contractPar \Trans^\OptPiOne)^{-1}(\PerTrans^\OptPiOne -
  \Trans^\OptPiOne)\qvalues} \\
&  = \contractPar
\opnorm{(\Id - \contractPar \PerTrans^\OptPiOne)^{-1}
  (\PerTrans^\OptPiOne - \Trans^\OptPiOne)} \cdot \inftynorm{(\Id -
  \contractPar \Trans^\OptPiOne)^{-1}(\PerTrans^\OptPiOne -
  \Trans^\OptPiOne)\qvalues} \\
& \leq \frac{\contractPar}{1 - \contractPar}
\opnorm{\PerTrans^\OptPiOne - \Trans^\OptPiOne} \cdot \inftynorm{(\Id
  - \contractPar \Trans^\OptPiOne)^{-1}(\PerTrans^\OptPiOne -
  \Trans^\OptPiOne)\qvalues} \\
& \leq \frac{\contractPar}{2} \inftynorm{(\Id - \contractPar
  \Trans^\OptPiOne)^{-1} (\bar{\Trans}^\OptPiOne -
  \Trans^\OptPiOne)\qvalues},
\end{align*}
where the last inequality uses
Lemma~\ref{lem:TransMat-properties}(b) and the first part of the
minimum sample size
assumption~\eqref{eqn:min-sample-size-requirement}. Combining this result with the bound~\eqref{eqn:Q-value-Lower-bound-1} we conclude
\begin{align*}
\inftynorm{\bar{\qvalues} - \qvalues} \geq \frac{\contractPar}{2} \inftynorm{(\Id - \contractPar \Trans^\OptPiOne)^{-1} (\bar{\Trans}^\OptPiOne - \Trans^\OptPiOne) \qvalues}. 
\end{align*}
With this result in hand, substituting the value of the transition kernel 
$\PmatBar$ from equation~\eqref{eqn:Perturbed-Transition} and recalling the 
definition of state-action pair $\sapair$ from equation~\eqref{eqn:shorthands-transition-mat} we have
\begin{align*}
\sqrt{\Qnumobs} \cdot \inftynorm{\bar{\qvalues} - \qvalues} \geq \frac{\discount \sqrt{\Qnumobs}}{2} \cdot (\Id - \contractPar \Trans^\OptPiOne)^{-1}(\PerTrans^\OptPiOne - \Trans^\OptPiOne) \qvalues(\optsapair) 
&= \frac{\discount\sqrt{\Qnumobs}}{2\sqrt{2}} \cdot \sum_{\sapair} \InvTermInd{\optsapair}{\sapair} \cdot (\PerTrans^\OptPiOne - \Trans^\OptPiOne) \qvalues(\sapair) \\
&\stackrel{(i)}{\geq} \frac{\discount}{4 \CompTermEntry(\optsapair)} \sum_{\sapair} (\InvTermInd{\optsapair}{\sapair})^2 \SigTerm^2(\sapair) \\
&\stackrel{(ii)}{=} \frac{\discount \CompTermEntry(\optsapair)}{4} = \frac{1}{4} \cdot 
 \max_{\policy \in \OptpolicySet} \inftynorm{ \contractPar \CompTerm(\policy; \Trans, \reward)},
\end{align*}
where step (i) follows by substituting the value of the transition kernel $\PmatBar$ (cf.~Proof of Lemma~\ref{lem:TransMat-properties} part (c)), and step (ii) follows by using the expression~\eqref{eqn:target-var-expression}.   
This completes the proof of part (a) of Lemma~\ref{lem:Q-function-LB}.

\subsubsection{Proof of Lemma~\ref{lem:Q-function-LB}(b)}

Borrowing the notation from part (a) of the proof, let $\sapair$
denote a generic element of the state-action set $\stateset \times
\actionset$.  Let $\OptPiTwo \in \arg\max_{\policy \in \OptpolicySet}
\inftynorm{\RewardTerm(\policy; \Trans, \reward)}$.  We use the
shorthands
\begin{align}
\label{eqn:shorthands-reward}  
\RewardTerm^2(\optsapair) \defn \max_{\policy \in \OptpolicySet}
\inftynorm{\RewardTerm(\policy; \Trans, \reward)}^2 =
\inftynorm{\RewardTerm(\OptPiTwo; \Trans, \reward)}^2, \qquad
\text{and} \quad \InvTerm \defn (\Id -
\contractPar\Trans^\OptPiTwo)^{-1}.
\end{align}
We define our perturbed reward function to be
\begin{align}
\label{eqn:perturbed-reward}
\Perreward(\sapair) = \reward(\sapair) +
\frac{1}{\RewardTerm(\optsapair) \sqrt{2\Qnumobs}}
\InvTermInd{\optsapair}{\sapair} \sgPar_\reward^2 \quad \text{for}
\quad z \in \stateset \times \actionset.
\end{align}
For $\probinstance_2 \defn (\Pmat, \Perreward)$, a short computation
shows that the Hellinger distance between the components of the instance pair
$(\probinstance, \probinstance_2)$ takes the form
\begin{align*}
  \dhel{\probinstance}{\probinstance_2}^2 \leq D_{KL}( \Ncal(\reward,
  \sigma_r^2 \Id) \mid \Ncal(\Perreward, \sigma_r^2\Id)) =
  \frac{1}{2\sigma_r^2} \twonorm{\reward - \Perreward}^2.
\end{align*}
Substituting the value of the reward $\Perreward$ from
equation~\eqref{eqn:perturbed-reward} yields
\begin{align*}
\dhel{\probinstance}{\probinstance_2}^2 \leq
\frac{1}{2\sigma_r^2}\twonorm{\Perreward - \reward}^2 =
\frac{1}{\RewardTerm^2(\optsapair) \cdot 4 \Qnumobs} \sum_{\sapair}
(\InvTermInd{\optsapair}{\sapair})^2 \sgPar_\reward^2 =
\frac{1}{4\Qnumobs},
\end{align*}
where the last equality uses the definition of the term
$\RewardTerm^2(\optsapair)$, i.e.,
\begin{align}
\label{eqn:reward-var-relation}
\RewardTerm^2(\optsapair) = \sum_{\sapair'}
(\InvTermInd{\optsapair}{\sapair'})^2 \sgPar_\reward^2.
\end{align}
It remains to prove a lower bound on the $\ell_\infty$-norm between the
optimal $Q$-functions for instances $\probinstance$ and
$\probinstance_2$.

Let $\qvalues \defn \qvalues(\Trans, \reward)$, and $\Perqvalues \defn
\qvalues(\Trans, \Perreward)$ be the optimal $Q$ functions for MDP
instances \mbox{$\probinstance \defn (\Trans, \reward)$} and
$\probinstance_2 \defn (\Trans, \Perreward)$, respectively.  Note that
$\qvalues$ and $\Perqvalues$ satisfy the Bellman equations
\begin{align}
\label{eqn:reward-Bellman}
\qvalues = \reward + \contractPar \Trans^\OptPiTwo \qvalues, \quad
\mbox{and} \quad \Perqvalues = \Perreward + \contractPar
\Trans^{\Perpolicy} \Perqvalues,
\end{align}
where $\Perpolicy$ is an optimal policy for the MDP instance $(\Trans,
\Perreward)$. By the optimality of policy $\Perpolicy$, we have the
entrywise inequality \mbox{$\Trans^{\Perpolicy} \Perqvalues \succeq
  \Trans^\OptPiTwo \Perqvalues$}; as a result, we have
\begin{align*}
(\Id - \contractPar \Trans^\OptPiTwo) \Perqvalues \succeq \Perreward \implies \Perqvalues \succeq (\Id - \contractPar \Trans^\OptPiTwo)^{-1} \Perreward,
\end{align*}
where the last step uses the fact that $(\Id - \contractPar \Trans^\OptPiTwo)^{-1}$ is entry-wise non-negative.
Combining the last inequality with the Bellman equation~\eqref{eqn:reward-Bellman} we have that

\begin{align}
\Perqvalues - \qvalues \succeq (\Id - \contractPar \Trans^\OptPiTwo)^{-1} (\Perreward - \reward) 
\end{align}
and that
\begin{align*}
\|(\Id - \contractPar \Trans^\OptPiTwo)^{-1} (\Perreward - \reward)\|_\infty \geq  (\Id - \contractPar \Trans^\OptPiTwo)^{-1} (\Perreward - \reward)(\optsapair) &= \frac{1}{\RewardTerm(\optsapair) \sqrt{2\Qnumobs}} \sum_\sapair (\InvTermInd{\optsapair}{\sapair})^2 \sgPar_\reward^2 \\
&= \frac{\RewardTerm(\optsapair)}{\sqrt{2\Qnumobs}},
\end{align*}
where the last equality uses the
relation~\eqref{eqn:reward-var-relation}.  Putting together the
pieces, we have shown that
\begin{align*}
  \inftynorm{\Perqvalues - \qvalues} \geq \frac{\RewardTerm(\optsapair)}{\sqrt{2\Qnumobs}} = \frac{1}{\sqrt{2\Qnumobs}} \cdot \max_{\policy \in \OptpolicySet} \inftynorm{\RewardTerm(\policy; \Trans, \reward)},
\end{align*}
as desired.

\section{Proof of Theorem~\ref{ThmUpper}}
\label{SecProofUpper}

In the section, we provide a proof of the upper bounds stated in
Theorem~\ref{ThmUpper}. Throughout the proof, we adopt the following
shorthands
\begin{gather}
\SpanTerm = \frac{\VecSpan{\Qstar}}{(1 - \contractPar)} \cdot
\log(8\Dim\NumEpoch/\pardelta), \qquad \notag \TauOpt = \max_{\pistar
  \in \OptPolicySet} \infnorm{\Nu(\pistar ; \instance)} \cdot
\sqrt{\log(8\Dim\NumEpoch/\pardelta)}, \\ \qquad \text{and} \qquad
\TauMax = \frac{1 + \infnorm{\reward} + \sigma_\reward \sqrt{1 -
    \contractPar}}{(1 - \gamma)^{1.5}} \cdot
\sqrt{\log(8\Dim\NumEpoch/\pardelta)}.
 \label{proof-shorthands}
\end{gather}
%


\subsection{Proof of Theorem~\ref{ThmUpper}(a)}

Our proof is based on the following two lemmas characterizing the behavior
of~\ref{AlgVR} across epochs.
\begin{lemma}
\label{lem:EpochLemma-small}
Under the assumptions of Theorem~\ref{ThmUpper}, for
each epoch $\epoch = 1, \ldots, \epochs$, we have
\begin{align}
\label{eqn:EpochWorstCaseBound}
\inftynorm{\Qbar_{\epoch + 1} - \Qstar} \leq
\frac{\inftynorm{\Qbar_\epoch - \Qstar}}{16} + c \left(
\frac{\TauMax}{\sqrt{\Nm}} + \frac{\SpanTerm}{\Nm} \right),
\end{align}
with probability at least $1 - \frac{\pardelta}{\epochs}$.
\end{lemma}

\noindent Lemma~\ref{lem:EpochLemma-small} follows by an argument
similar to that used in the proof of Theorem~1 of the
paper~\cite{wainwright2019variancereduced}, so we omit the details
here.  See also the proof of Lemma~\ref{lem:EpochLemma-large} for some
relevant arguments.
\begin{lemma}
\label{lem:EpochLemma-large}
Under the assumptions of Theorem~\ref{ThmUpper}, for
epochs $\epoch$ such that the re-centering sample size $\Nm$ satisfies
the bound $\Nm \geq \LogDM \frac{(1 + \inftynorm{r} + \sigma_\reward
  \sqrt{1 - \contractPar})^2}{\SpGap^2(1 - \contractPar)^3}$, we have
\begin{align}
\label{eqn:SharpCorBound}
\infnorm{\Qbar_{m + 1} - \Qstar} \leq \frac{\infnorm{\Qbar_{m} -
    \Qstar}}{16} + c \cdot \left( \frac{\TauOpt}{\sqrt{\Nm}} +
\frac{\SpanTerm}{\Nm} \right),
\end{align}
with probability at least $1 - \frac{\pardelta}{\epochs}$. 
\end{lemma}
\noindent See Section~\ref{proof-of-lemma:EpochLemma-large} for the
proof of Lemma~\ref{lem:EpochLemma-large}. 

\subsubsection{Completing the proof}

Using the two lemmas above, we can now complete the proof of
Theorem~\ref{ThmUpper}(a).  Recalling the epoch sample size
formula~\eqref{EqnPERecenterSize-smallsample}, we see that the
bound~\eqref{eqn:SharpCorBound} holds for all epochs
\begin{align*}
  \epoch \geq \epochstar \defn \log_2 \frac{1 + \inftynorm{r} +
    \sigma_\reward \sqrt{1 - \contractPar}}{\SpGap \sqrt{1 -
      \contractPar}}.
\end{align*}
Observe that the minimum sample size requirement from
Theorem~\ref{ThmUpper} ensures that $\epochs \geq
\epochstar$.  Now, applying the recursions~\eqref{eqn:SharpCorBound}
and~\eqref{eqn:EpochWorstCaseBound} we obtain
\begin{align*}
  \inftynorm{\Qbar_{\epochs + 1} - \Qstar} & \leq \frac{\inftynorm{
      \Qbar_{\epochs} - \Qstar}}{16} + c\left(
  \frac{\TauOpt}{\sqrt{\NM}} + \frac{\SpanTerm}{\NM}\right)
  \\ &\stackrel{(i)}{\leq} \frac{\inftynorm{\Qbar_{\epochstar} -
      \Qstar}}{16^{\epochs - \epochstar}} + c \cdot \left( \sum_{k =
    0}^{\epochs - \epochstar}
  \frac{\TauOpt}{16^{k}\sqrt{\Qnumobs_{\epochs - k}}} +
  \frac{\SpanTerm}{ 16^k \cdot \Qnumobs_{\epochs - k}} \right) \\
  & \stackrel{(ii)}{\leq} \frac{\inftynorm{\Qbar_{1} -
      \Qstar}}{16^{\epochs}} + c \cdot \left( \sum_{k = 0}^{\epochs -
    \epochstar} \frac{\TauOpt}{16^{k}\sqrt{\Qnumobs_{\epochs - k}}} +
  \sum_{k=\epochs - \epochstar + 1}^{\epochs}
  \frac{\TauMax}{16^{k}\sqrt{\Qnumobs_{\epochs - k}}} \right) \\ \\
   & \qquad \qquad \qquad \qquad \qquad \qquad\qquad\qquad\qquad  +  c \cdot  \sum_{k = 0}^{\epochs} \frac{\SpanTerm}{16^k \Qnumobs_{\epochs - k}}   \\
   & \stackrel{(iii)}{\leq} \frac{\inftynorm{\Qbar_{1} - \Qstar}}{16^{\epochs}}
   + c \cdot \left( \frac{\TauMax}{8^{\epochs - \epochstar} \cdot \sqrt{\Qnumobs_{\epochs}}}
   + \frac{\TauOpt}{\sqrt{\Qnumobs_\epochs}} + \frac{\SpanTerm}{\Qnumobs_\epochs}  \right).
\end{align*}
Inequality (i) follows via repeated application of the
recursion~\eqref{eqn:SharpCorBound}, inequality (ii) follows via
repeated application of the recursion~\eqref{eqn:EpochWorstCaseBound},
and inequality (iii) utilizes the relation ${\Qnumobs_{\epochs - k}
  \cdot 4^{k} = \Qnumobs_\epochs}$
(cf. definition~\eqref{EqnPERecenterSize-smallsample}).  Via the union
bound, the above inequalities hold simultaneously with probability at least $1
- \delta$.  Next, note that by our choice of $\Nm$, we have the
inequality $2 \Qnumobs_\epochs \leq \Qnumobs \leq \frac{8}{3}
\Qnumobs_\epochs$. Putting together the pieces, we conclude that
\begin{multline}
  \inftynorm{\Qbar_{\epochs + 1} - \Qstar} \leq c \cdot
  \inftynorm{\Qbar_1 - \Qstar} \cdot
  \frac{\log^2((8\numstates/\pardelta) \cdot \log \numobs) }{\numobs^2
    (1 - \contractPar)^4} + c \cdot \frac{(1 + \inftynorm{\reward} +
    \sigma_\reward \sqrt{1 - \contractPar})^4 }{(1 -
    \gamma)^{1.5}\sqrt{\numobs}} \cdot
  \frac{\log^2((8\numstates/\pardelta) \cdot \log
    \numobs)}{\Qnumobs^{3/2} (1 - \contractPar)^{\frac{9}{2}}
    \SpGap^3} \\
     + c \cdot \left(  \sqrt{\frac{\LogDM}{\Qnumobs}} \CompTermInstance 
        + \frac{\LogDM}{\Qnumobs} \cdot \frac{\VecSpan{\Qstar}}{1 - \contractPar}  \right).
\label{eqn:instance-ub-all-terms} 
\end{multline}
Substituting the lower bound condition 
\begin{align*}
\frac{\Qnumobs}{\log^2(\Qnumobs)} \geq c\log(\numstates / \pardelta) \frac{(1 + \inftynorm{\reward} + \sigma_\reward \sqrt{1 - \contractPar})^2}{(1 - \contractPar)^3} \cdot \max \left\{1,  \frac{1}{\SpGap^2 \cdot (1 - \contractPar)^\beta} \right\}
\end{align*}
yields the claimed bound. All that remains is to verify the choice of
batch sizes $\{ \Nm \}_{\epochs = 1}^\epochs$ is a valid choice, i.e.,
we need to verify that the algorithm~\ref{AlgVR} with parameter
choices~\eqref{EqnQParam-small-sample} uses at most $\Qnumobs$
samples.  Recall that the total number of samples used in the
$\epochs$ epochs is given by $K \epochs + \sum_{\epoch = 1}^\epochs
\Nm$. Substituting the values of $\Nm$ and $\epochs$ from
equations~\eqref{EqnQParam-small-sample} we obtain
\begin{align*}
K\epochs +  \sum_{\epochs = 1}^M \Nm &\leq c \cdot \LogDM \cdot \sum_{\epoch = 1}^\epochs \frac{4^\epoch}{(1 - \contractPar)^2} + \frac{\Qnumobs}{2} \\
&\leq c' \cdot \LogDM \cdot \frac{4^\epochs}{(1 - \contractPar)^2} \leq \frac{\Qnumobs}{2} + \frac{\Qnumobs}{2} \leq \Qnumobs.
\end{align*}
This completes the proof of
Theorem~\ref{ThmUpper}(a).

\paragraph{Comment on the lower-order terms:}

Here, we argue that the first two terms in the right-hand side of the
bound~\eqref{eqn:instance-ub-all-terms} are of lower order.  A careful
look at the proof reveals that for any $p \geq 1$ by increasing our
choice of $\Nm$ by a constant factor depending on $p$, we can bound
the first term by
\begin{align*}
c_1 \cdot \frac{\inftynorm{\Qbar_1 - \Qstar}}{\Qnumobs ^ p} \cdot
\frac{\log^p((8 \numstates / \pardelta) \cdot \log \Qnumobs)}{(1 -
  \contractPar)^{2p}},
\end{align*}
and the second term by
\begin{align*}
c_2 \cdot \frac{(1 + \inftynorm{r} + \sigma_\reward \sqrt{1 -
    \contractPar})^{3q + 1}}{(1 - \contractPar)^{1.5}\sqrt{\Qnumobs}} \cdot
\frac{\log^{2q}(( 8 \numstates / \pardelta) \cdot \log
  \Qnumobs)}{\Qnumobs^{3q/2} (1 - \contractPar)^{\frac{9q}{2}}
  \SpGap^{3q}},
\end{align*}
where $q = \frac{2}{3} p - \frac{1}{3}$, and $(c_1, c_2)$ are
universal constants only depending on $(p, q)$. The number of samples
satisfies $\Qnumobs \gtrsim \frac{(1 + \inftynorm{\reward} +
  \sigma_\reward \sqrt{1 - \contractPar})^2}{\SpGap^2 (1 -
  \contractPar)^{3 + \beta}}$ by assumption, and consequently, the two
terms can be made arbitrarily small by increasing $(p, q)$
appropriately.  The equation~\eqref{eqn:instance-ub-all-terms}
displays the bound for the pair $(p, q) = (2,1)$. \\

\noindent The only remaining detail is to prove
Lemma~\ref{lem:EpochLemma-large}.


\subsubsection{Proof of Lemma~\ref{lem:EpochLemma-large}}
\label{proof-of-lemma:EpochLemma-large}
Recall that the update within an epoch takes the form
(cf.~\ref{AlgRunEpoch})
\begin{align*}
\Q_{k+1} = (1 - \stepsize_k) \Q_k + \stepsize_k \left \lbrace
\NoisyOp_k(\Q) - \NoisyOp_k(\Qbar_m) + \RecenterOp(\Qbar_m) \right
\rbrace,
\end{align*}
where $\Qbar_m$ represents the input into epoch $\epoch$. We define
the shifted operators and noisy shifted operators for epoch $\epoch$
by
\begin{align}
\label{eqn:shifted-bellman}
\shiftedOp(\Q) = \Op(\Q) - \Op(\Qbar_m) + \RecenterOp(\Qbar_m) \quad
\text{and} \quad \NoisyShiftedOp_k(\Q) = \NoisyOp_k(\Q) -
\NoisyOp_k(\Qbar_m) + \RecenterOp(\Qbar_m).
\end{align}
Since both of the operators $\Op$ and $\NoisyOp_k$ are
$\contractPar$-contractive in the $\ell_\infty$-norm, the operators
$\shiftedOp$ and $\NoisyShiftedOp_k$ are also
$\contractPar$-contractive operators in the same norm.  Let
$\Qhat_\epoch$ denote the unique fixed point of the operator $\shiftedOp$. The
roadmap of the proof is to show that at the end of epoch $\epoch$, the
estimate $\Q_{K + 1}$ is close to the fixed point $\Qhat_\epoch$ for
sufficiently large value of the epoch length $K$ and that the fixed
point $\Qhat_\epoch$ is closer to $\Qstar$ than the epoch
initialization $\Qbar_m$ for sufficiently large $\Nm$.

The proof of Lemma~\ref{lem:EpochLemma-large} relies on two auxiliary
lemmas that formalize this intuition. Lemma~\ref{lem:gen-SA-bound}
characterizes the progress of Algorithm~\ref{AlgVR} within an epoch,
and Lemma~\ref{lem:Instance-Dependent-large} addresses the progress
of Algorithm~\ref{AlgVR} over the epochs.
\begin{lemma}
\label{lem:gen-SA-bound}
Given an epoch length $\EpLen$ lower bounded as $\EpLen \geq c_2
\frac{\log(\Qnumobs \numstates / \pardelta)}{(1 - \contractPar)^3}$,
we have
\begin{align*}
  \infnorm{\Q_{K + 1} - \Qhat_\epoch} 
  \leq \frac{1}{33}  \infnorm{\Qbar_\epoch - \Qstar}
  + \frac{1}{33} \infnorm{\Qhat_m - \Qstar},
\end{align*}
with probability exceeding $1 - \frac{\pardelta}{2\NumEpoch}$.
\end{lemma}
\noindent
Lemma~\ref{lem:gen-SA-bound} is borrowed from the paper~\cite{khamaru2020PE}; see the proof of Lemma 2 in that paper for details.

%
%

Our next lemma bounds the difference between the epoch fixed point
$\Qhat_\epoch$ and the optimal value function $\Qstar$.
\begin{lemma}
\label{lem:Instance-Dependent-large}
Assume that $\Nm$ satisfies the bound \mbox{$ \Nm \geq c \LogDM
  \frac{(1 + \infnorm{\reward} + \sigma_r\sqrt{1 -
      \contractPar})^2}{\SpGap^2(1 - \contractPar)^3}$}. Then we have
\begin{align*}
\inftynorm{\Qhat_\epoch - \Qstar} \leq \frac{\inftynorm{\Qbar_\epoch -
    \Qstar}}{33} + c_4 \left\{ \frac{\TauOpt}{\sqrt{\Nm}} +
\frac{\SpanTerm}{\Nm} \right\},
\end{align*}
with probability exceeding $1 - \frac{\pardelta}{2\epochs}$.
\end{lemma}
\noindent See Appendix~\ref{Sec:ProofEpochLemma} for a proof of this
lemma. \\

With these two auxiliary results in hand, completing the proof of
Lemma~\ref{lem:EpochLemma-large} is relatively straightforward.  By
the triangle inequality, we have
\begin{align}
\infnorm{\Qbar_{\epoch + 1} - \Qstar} \equiv \infnorm{\Q_{K + 1} -
  \Qstar} & \leq \infnorm{\Q_{K + 1} - \Qhat_\epoch} +
\infnorm{\Qhat_\epoch - \Qstar} \notag \\
& \stackrel{(i)}{\leq} \left \{ \frac{1}{32} \infnorm{\Qbar_\epoch -
  \Qstar} + \frac{1}{32} \infnorm{\Qhat_m - \Qstar} \right \} +
\infnorm{\Qhat_\epoch - \Qstar} \notag \\ & = \frac{1}{32}
\infnorm{\Qbar_\epoch - \Qstar} + \frac{33}{32} \infnorm{\Qhat_m -
  \Qstar} \notag \\
& \stackrel{(ii)}{\leq} \frac{1}{32} \infnorm{\Qbar_\epoch - \Qstar} +
\frac{1}{32} \left\lbrace \infnorm{\Qbar_\epoch - \Qstar}
\right\rbrace + c\left(\frac{\TauOpt}{\sqrt{\Nm}} +
\frac{\SpanTerm}{\Nm} \right) \notag \\
\label{eqn:gen-main-recursion} .
& \leq \frac{1}{16} \infnorm{\Qbar_\epoch - \Qstar} + +
c\left(\frac{\TauOpt}{\sqrt{\Nm}} + \frac{\SpanTerm}{\Nm} \right).
\end{align}
Here inequality (i) follows from Lemma~\ref{lem:gen-SA-bound}, whereas
inequality (ii) follows from Lemma~\ref{lem:Instance-Dependent-large}.
Finally, the two bounds hold jointly with probability at least $1 -
\frac{\pardelta}{\NumEpoch}$ via a union bound.


\subsection{Proof of Theorem~\ref{ThmUpper}(b)}

This argument follows the same structure as the proof of part (a) of
Theorem~\ref{ThmUpper}; we retain the same shorthands from
equation~\eqref{proof-shorthands}.  Our proof uses
Lemma~\ref{lem:EpochLemma-small} along with the following modification
of Lemma~\ref{lem:EpochLemma-large}.
\begin{lemma}
\label{lem:EpochLemma-instance-gen}
Under the conditions of Theorem~\ref{ThmUpper}(b), for epochs $\epoch$
such that the re-centering sample size $\Nm$ satisfies the bound $\Nm
\geq \LogDM \frac{L^2 (1 + \inftynorm{r} + \sigma_\reward \sqrt{1 -
    \contractPar})^2}{(1 - \contractPar)^5}$, we have
\begin{align}
\inftynorm{\Qbar_{\epoch + 1} - \Qstar} \leq \frac{\inftynorm{\Qbar_\epoch - \Qstar}}{16} + c \cdot \left( \frac{\TauOpt}{\sqrt{\Nm}} + \frac{\SpanTerm}{\Nm} \right),
\end{align}
with probability at least $1 - \frac{\pardelta}{\epochs}$.
\end{lemma}
\noindent
See Appendix~\ref{Sec:ProofGenEpochLemma} for a proof of this lemma. \\

Observe that Lemma~\ref{lem:EpochLemma-instance-gen} holds for all
epochs $\epoch \geq \epochstar \defn \log_2 \frac{L(1 +
  \inftynorm{\reward} + \sigma_\reward\sqrt{1 - \contractPar})}{(1 -
  \contractPar)^{3/2}}$.  Invoking Lemma~\ref{lem:EpochLemma-small}
for all $\epoch < \epochstar$ and
Lemma~\ref{lem:EpochLemma-instance-gen} for all epochs $\epoch \geq
\epochstar$, and doing a calculation similar to the proof of part (a),
yields
\begin{align*}
\inftynorm{\Qbar_{\epochs + 1} - \Qstar} \leq \frac{\inftynorm{\Qbar_1
    - \Qstar}}{16^\epochs} + c \cdot \left(\frac{\TauMax}{8^{\epochs -
    \epochstar} \sqrt{\NM}} + \frac{\TauOpt}{\sqrt{\NM}} +
\frac{\SpanTerm}{\NM} \right),
\end{align*}
with probability exceeding $1 - \pardelta$. Finally, our choice of the
epoch size $\Nm$
(cf. definition~\eqref{EqnPERecenterSize-smallsample}) ensures $2\NM
\leq \Qnumobs \leq \frac{8}{3} \NM$, and substituting the values of
the triple $(\Nm, \epochstar, \epochs)$ we conclude that
\begin{align*}
\inftynorm{\Qbar_{\epochs + 1} - \Qstar} &\leq c \cdot
\inftynorm{\Qbar_1 - \Qstar} \cdot \frac{\log^2((8\numstates /
  \pardelta) \cdot \log \Qnumobs)}{\Qnumobs^2 ( 1 - \contractPar)^4} +
c \cdot \frac{L^3 (1 + \inftynorm{\reward} + \sigma_\reward \sqrt{1 -
    \contractPar})^4}{(1 - \contractPar)^{1.5}\sqrt{\Qnumobs}} \cdot
\frac{\log^2((8\numstates / \pardelta) \cdot \log
  \Qnumobs)}{\Qnumobs^{1.5} (1 - \contractPar)^{7.5}} \\ & + c \cdot
\left( \sqrt{\frac{\LogDM}{\Qnumobs}} \CompTermInstance +
\frac{\LogDM}{\Qnumobs} \cdot \frac{\VecSpan{\Qstar}}{1 -
  \contractPar} \right),
\end{align*}
\paragraph{Comment on the lower-order terms:} For any $p \geq 1$
by increasing our choice of $\Nm$ by a constant factor depending on
$p$, we can bound the first term via
\begin{align*}
c_1 \cdot \frac{\inftynorm{\Qbar_1 - \Qstar}}{\Qnumobs ^ p} \cdot
\frac{\log^p((8 \numstates / \pardelta) \cdot \log \Qnumobs)}{(1 -
  \contractPar)^{2p}},
\end{align*}
and the second term by
\begin{align*}
c_2 \cdot \frac{L^{3q} (1 + \inftynorm{\reward} + \sigma_\reward
  \sqrt{1 - \contractPar})^{3 q + 1}}{(1 - \contractPar)^{1.5}
  \sqrt{\Qnumobs}} \cdot \frac{\log^{3q/2}((8\numstates / \pardelta)
  \cdot \log \Qnumobs)}{\Qnumobs^{3q / 2} ( 1 -
  \contractPar)^{15q/2}},
\end{align*}
with $q = \frac{2}{3} p - \frac{1}{3}$. The number of samples
satisfies $\Qnumobs \gtrsim \frac{L^2 (1 + \inftynorm{\reward} +
  \sigma_\reward\sqrt{1- \contractPar})^2}{(1 - \contractPar)^{5 +
    \beta}}$ by assumption, so the two can be made arbitrarily small
by increasing $(p, q)$ appropriately.  This completes the proof of
Theorem~\ref{ThmUpper}(b).


\section{Discussion}
\label{SecDiscussion}


In this paper, we presented an analysis of $Q$-learning through the
instance-dependent framework in the synchronous setting. For
$\gamma$-discounted MDPs with finite state space $\mathcal{X}$ and
action space $\mathcal{U}$, we have proved a local non-asymptotic
lower bound for estimating the $Q$-function dependent on a functional
$\AllCompTerm(\cdot; \TranMatQ, \reward)$ of the MDP instance
$(\TranMatQ, \reward)$ that measures the variance of solving for the
$Q$-function. In addition, we have provided an analysis of a form of
variance-reduced $Q$-learning and obtained instance-dependent
guarantees on the $\ell_\infty$-error for sample sizes
\mbox{$\frac{\Qnumobs}{(\log \Qnumobs)^2} \geq c \cdot
  \frac{\log(\numstates / \pardelta)}{(1 - \gamma)^3 \SpGap^@}$} and
\mbox{$\frac{\Qnumobs}{(\log \Qnumobs)^2} \geq c \frac{\log(\numstates
    / \pardelta)}{(1 - \contractPar)^5}$} on Lipschitz MDPs that match
the corresponding lower bound, establishing instance-optimality. We
conjecture that optimality of Algorithm~\ref{AlgVR} still remains true
for general MDPs and sample sizes $\Qnumobs \geq \frac{\log(\numstates
  / \pardelta)}{(1 - \contractPar)^3}$, and is left for further
endeavours.





\appendix

\section{Calculations for Example~\ref{ExaSimple}}
\label{sec:simple-problem-details}

Here we derive the bound~\eqref{eqn:simple-instance-bound}.  Letting
$\Vstar$ denote the value function of the optimal policy
$\policystar$, we have
\begin{align}
 (\obsmat^{\policystar} - \probmat^{\policystar}) \Q = \begin{bmatrix}
    | & | \\ (\obsmat_{\action_1} - \probmat_{\action_1}) \Vstar & 0
    \\ | & |\end{bmatrix}.
\end{align}
Letting $\Wmat = (\Id - \contractPar \probmat_{\action_1})^{-1}
(\obsmat_{\action_1} - \probmat_{\action_1}) \Q_{\policystar}$ and
solving for $(\Id - \contractPar \probmat^{\policystar}) \randmat =
\contractPar (\obsmat^{\policystar} - \probmat^{\policystar}) \Q $
gives
\begin{align}
    \randmat = \contractPar \cdot \begin{bmatrix}
    | & | \\
     \Wmat & \contractPar \Wmat \\
    | & | 
    \end{bmatrix}.
\end{align}
Thus, we have
\begin{align*}
 \infnorm{\Nu(\pistar; \lambdaInstance)} & \defn \max_{(\state,
   \action)} \; \abs{\Nu(\pistar; \lambdaInstance)(\state, \action)} =
 \max_{(\state, \action)} \; \abs{ \sqrt{\var(\randmat) (\state,
     \action)}} \leq c \cdot \frac{1}{(1 - \contractPar)^{1.5 -
     \lambda}}
\end{align*}
The second equality above follows from the definition~\eqref{EqnKey}
of the matrix $\Nu(\pistar; \lambdaInstance)$, and the last step via
some simple calculations.

\section{Auxiliary lemmas for Theorem~\ref{ThmLower}}

In this section, we prove the auxiliary lemmas that are the used in
the proof of Theorem~\ref{ThmLower}.

\subsection{Proof of Lemma~\ref{lem:minimax-to-Hellinger}}
\label{proof:lem:minimax-to-Hellinger}
This proof uses standard arguments, in particular following the usual
avenue of reducing estimation to
testing~\citep{birge1983,dinosaur2019}. For completeness, we provide
the details here. We use $\qvalues$ and $\qvalues'$ to denote the
optimal $Q$-functions for problem $\probinstance$ and $\probinstance'$
respectively. We first lower bound the minimax risk over
$\probinstance, \probinstance'$ by the averaged risk as follows:
\begin{align*}
\inf_{\qestimate} \max_{\AltProb \in \{ \probinstance, \probinstance'
  \}} \Exs_\probdistr \left[ \inftynorm{\qvalues -
    \qvalues(\AltProb)}\right] \geq \frac{1}{2}
\left(\Exs_{\probdistr^\Qnumobs}\left[ \inftynorm{\qestimate -
    \qvalues} \right] + \Exs_{(\probdistr')^\Qnumobs}
\left[\inftynorm{\qestimate - \qvalues'}\right] \right).
\end{align*}
Here $\probdistr^\Qnumobs$ is a product measure that yields $\Qnumobs$ i.i.d. samples from
$\probinstance$. Then, for any $\pardelta \geq 0$, we have by Markov's
inequality
\begin{align*}
\Exs_{\probdistr^\Qnumobs}\left[ \inftynorm{\qestimate - \qvalues}
  \right] + \Exs_{(\probdistr')^\Qnumobs} \left[ \inftynorm{\qestimate
    - \qvalues'} \right] \geq \delta \left[ \probdistr^\Qnumobs \left(
  \inftynorm{\qestimate - \qvalues} \geq \pardelta \right) +
  (\probdistr')^\Qnumobs \left( \inftynorm{\qestimate - \qvalues'}
  \geq \delta \right) \right].
\end{align*}
Define $\pardelta_{01} \defn \frac{1}{2} \inftynorm{\qvalues -
  \qvalues'}$, we have
\begin{align*}
\inftynorm{\qestimate - \qvalues} < \pardelta_{01} \implies
\inftynorm{\qestimate - \qvalues'} > \pardelta_{01},
\end{align*}
yielding
\begin{align*}
\Exs_{\probdistr^\Qnumobs}\left[ \inftynorm{\qestimate - \qvalues}
  \right] + \Exs_{(\probdistr')^\Qnumobs} \left[ \inftynorm{\qestimate
    - \qvalues} \right] &\geq \pardelta_{01} \left[1 -
  \probdistr^\Qnumobs \left( \inftynorm{\qestimate - \qvalues} <
  \pardelta_{01} \right) + (\probdistr')^\Qnumobs\left(
  \inftynorm{\qestimate - \qvalues'} \geq \pardelta_{01}
  \right)\right] \\ &\geq \pardelta_{01} \left[1 - \probdistr^\Qnumobs
  \left( \inftynorm{\qestimate - \qvalues'} \geq \pardelta_{01}
  \right) + (\probdistr')^\Qnumobs\left( \inftynorm{\qestimate -
    \qvalues'} \geq \pardelta_{01} \right) \right] \\ &\geq
\pardelta_{01} \left[1 - \TVnorm{\probdistr^\Qnumobs -
    (\probdistr')^\Qnumobs} \right] \\ &\geq \pardelta_{01} \left[ 1 -
  \sqrt{2} \dhel{\probdistr^\Qnumobs}{(\probdistr')^\Qnumobs}^2
  \right].
\end{align*}
Via the tensorization property of the Hellinger distance for
independent random variables we have
\begin{align*}
\dhel{\probdistr^\Qnumobs}{(\probdistr')^\Qnumobs}^2 = 1 - \left(1 -
\dhel{\probdistr}{\probdistr'}^2 \right)^\Qnumobs \leq \Qnumobs
\dhel{\probdistr}{\probdistr'}^2.
\end{align*}
Putting together the pieces, we have that
\begin{align*}
\inf_{\qestimate} \max_{\AltProb \in \{ \probinstance,
  \probinstance'\}} \Exs_{\AltProb} \left[ \inftynorm{\qvalues -
    \qvalues(\AltProb)} \right] \geq \frac{1}{4}
\inftynorm{\qvalues(\probinstance) - \qvalues(\probinstance')} \cdot
\left(1 - \sqrt{2} \Qnumobs \cdot
\dhel{\probinstance}{\probinstance'}^2\right)_+.
\end{align*}
The desired result then follows from taking a supremum over all positive
alternative $\probinstance' \in \Alter$ and  a simple calculation.

\subsection{Proof of Lemma~\ref{lem:TransMat-properties}}
\label{subsec:proof-TransMat-properties}

We devote a subsection to each of the three parts of this lemma.


\subsubsection{Proof of Lemma~\ref{lem:TransMat-properties}(a)}

In order to establish that $\PmatBar$ is a transition kernel, we
observe that
\begin{align*}
\sum_{\state'} \PerTransInd{\sapair}{\state'} = 1 + \frac{1}{
  \CompTermEntry(\optsapair) \sqrt{2\Qnumobs} }
\InvTermInd{\optsapair}{\sapair} \cdot \left(\sum_{\state'}
\TransInd{\sapair}{\state'} (\qvalues(\state', \OptPiOne(\state')) -
(\Trans^\OptPiOne \qvalues)(\sapair))\right) = 1,
\end{align*}
where the last equality above follows by noting that
$(\Trans^\OptPiOne \qvalues)(\sapair) = \sum_{\state'}
\TransInd{\sapair}{\state'} \qvalues(\state', \OptPiOne(\state'))$.
To check non-negativity of entries of $\PmatBar$ note we have
$\abs{\InvTermInd{\sapair}{\sapair'}} \leq \frac{1}{1 -
  \contractPar}$, and \mbox{$2\VecSpan{\qvalues} \geq
  \abs{\qvalues(\state', \OptPiOne(\state')) - (\Trans^\OptPiOne
    \qvalues)(\sapair)}$}. Combining the last two observation along
with the sample size
requirement~\eqref{eqn:min-sample-size-requirement} implies
\begin{align*}
\PerTransInd{\sapair}{\state'} \geq 1 -
\frac{1}{\CompTermEntry(\optsapair) \sqrt{2\Qnumobs}} \cdot \frac{
  \|\theta\|_{\mathrm{span}} }{1 - \discount} \geq 0,
\end{align*}
establishing that $\PmatBar$ defines a valid set of transition
kernels.


\subsubsection{Proof of Lemma~\ref{lem:TransMat-properties}(b)}

The proof of part (b) follows by first providing a general upper bound
on the Hellinger distance $\dhel{\probinstance}{\probinstance_1}$, and
then substituting the values of instances $\probinstance$ and
$\probinstance_1$. Concretely, we prove
\begin{align}
\label{eqn:Hellinger-bound}
  \HelDist^2(\probinstance, \probinstance_1) \;\;
  \labelrel{\leq}{eqn:Hellinger-bound-1} \;\; \frac{1}{2} \cdot
  \sum_{\sapair, \state'} \frac{(\TransInd{\sapair}{\state'} -
    \PerTransInd{\sapair}{\state'})^2}{\TransInd{\sapair}{\state'}}
  \;\; \labelrel{\leq}{eqn:Hellinger-bound-2} \;\;
  \frac{1}{4\Qnumobs}.
\end{align}
With this result in hand, the claimed bound on
$\opnorm{\PerTrans^\OptPiOne - \Trans^\OptPiOne}$ is
immediate. Indeed,
\begin{align*}
  \opnorm{\PerTrans^\OptPiOne - \Trans^\OptPiOne}^2 \leq
  \sum_{\sapair, \state'} (\TransInd{\sapair}{\state'} -
  \PerTransInd{\sapair}{\state'})^2 \leq \sum_{\sapair, \state'}
  \frac{(\TransInd{\sapair}{\state'} -
    \PerTransInd{\sapair}{\state'})^2}{\TransInd{\sapair}{\state'}}
  \leq \frac{1}{2\Qnumobs}.
\end{align*}
It remains to prove the bounds~\eqref{eqn:Hellinger-bound}(a)
and~\eqref{eqn:Hellinger-bound}(b).


\paragraph{Proof of~(\ref{eqn:Hellinger-bound}\ref{eqn:Hellinger-bound-1}):}

We use $(\obsmat, \obsreward)$ (respectively $(\obsmat',
\obsreward')$) to denote a sample drawn from the distribution $\distr$
(respectively $\distr'$), and $\distr_\obsmat, \distr_\obsreward$
(respectively $\distr_\obsmat', \distr_\obsreward'$) to denote the
marginal distribution of $\obsmat, \obsreward$ (respectively
$\obsmat', \obsreward'$). By independence of $\obsmat$ and
$\obsreward$ (and likewise for $\obsmat', \obsreward'$) we have
\begin{align}
\distr = \distr_\obsmat \otimes \distr_\obsreward, \quad \text{and}
\quad \distr' = \distr_\obsmat' \otimes \distr_\obsreward'.
\end{align}
Let $\probinstance' = (\Trans', \reward') \in \Alter_1$ (so $\reward'
= \reward$). Via the independence between $\obsmat$ and $\obsreward$,
we have
\begin{align}
\HelDist^2(\distr, \distr') = \HelDist^2(\distr_\obsmat,
\distr_\obsmat').
\end{align}
For state-action pairs $(\state, \action)$, $\obsmat(\state, \action)$
are independent (and likewise for $\obsmat'$) so
\begin{align*}
\HelDist^2(\distr_\obsmat, \distr_\obsmat') = 1 - \prod_{\state,
  \action}\left(1 - \HelDist(\distr_{\obsmat(\state, \action)},
\distr_{\obsmat'(\state, \action)})\right)^2 \leq \sum_{\state,
  \action} \HelDist^2(\distr_{\obsmat(\state, \action)},
\distr_{\obsmat'(\state, \action)}).
\end{align*}
Note that $\obsmat(\state, \action)$ and $\obsmat'(\state, \action)$
have multinomial distribution with parameters $\Trans_\action(\cdot
\mid \state)$ and $\Trans_\action'(\cdot \mid \state)$
respectively. Therefore,
\begin{align*}
\HelDist^2(\distr_{\obsmat(\state, \action)}, \distr_{\obsmat'(\state,
  \action)}) \leq \frac{1}{2} D_{\chi^2}\left(\distr_{\obsmat'(\state,
  \action)} \| \distr_{\obsmat(\state, \action)} \right) = \frac{1}{2}
\sum_{\state'} \frac{(\TransInd{\sapair}{\state'} -
  \PerTransInd{\sapair}{\state'})^2}{\TransInd{\sapair}{\state'}}.
\end{align*}

\paragraph{Proof of~(\ref{eqn:Hellinger-bound}\ref{eqn:Hellinger-bound-2}):}
We have
\begin{align*}
\sum_{\sapair, \state'} \frac{(\TransInd{\sapair}{\state'} -
  \PerTransInd{\sapair}{\state'})^2}{\TransInd{\sapair}{\state'}} &=
\frac{1}{2\Qnumobs \CompTermEntry^2(\optsapair)} \sum_{\sapair}
\sum_{\state'} \TransInd{\sapair}{\state'}
(\InvTermInd{\optsapair}{\sapair})^2 (\qvalues(\state',
\OptPiOne(\state')) - (\Trans^\OptPiOne \qvalues)(\sapair))^2 \\
&= \frac{1}{2\Qnumobs \CompTermEntry^2(\optsapair)} \sum_{\sapair}
(\InvTermInd{\optsapair}{\sapair})^2 \cdot \left( \sum_{\state'}
\TransInd{\sapair}{\state'} (\qvalues(\state', \OptPiOne(\state')) -
(\Trans^\OptPiOne \qvalues)(\sapair))^2 \right) \\
&\stackrel{(i)}{=} \frac{1}{2\Qnumobs \CompTermEntry^2(\optsapair)}
\sum_\sapair (\InvTermInd{\optsapair}{\sapair})^2 \SigTerm^2(\sapair)
\\
&\stackrel{(ii)}{=} \frac{1}{2\Qnumobs},
\end{align*}
Equality (i) follows from the definition
\begin{align}
\label{eqn:psi-var-defn}
  \SigTerm^2(\sapair) = \text{Var}\left(\obsmat^\OptPiOne
  \qvalues(\sapair)\right) = \sum_{\state'}
  \TransInd{\sapair}{\state'} (\qvalues(\state', \OptPiOne(\state')) -
  (\Trans^\OptPiOne \qvalues)(\sapair))^2,
\end{align}
whereas equality (ii) follows from the
definition~\eqref{eqn:shorthands-transition-mat}, which ensures that
\begin{align*}
  \CompTermEntry^2(\optsapair) = \text{Var}\left((\Id - \contractPar
  \Trans^\OptPiOne)^{-1} \obsmat^\OptPiOne \qvalues(\optsapair)\right)
  = \sum_{\sapair'} (\InvTermInd{\sapair}{\sapair'})^2
  \SigTerm^2(\sapair').
\end{align*}


\subsubsection{Proof of Lemma~\ref{lem:TransMat-properties}(c)}

The entries of the matrix $\InvTerm \defn (\Id - \contractPar
\Trans^\OptPiOne)^{-1}$ are positive, so that it suffices to show that
the vector $(\PerTrans^\policy - \Trans^\policy) \qvalues$ is
entry-wise positive. We have
\begin{align*}
(\PerTrans^\OptPiOne - \Trans^\OptPiOne) \qvalues(\sapair) &=
  \sum_{\state'} (\PerTransInd{\sapair}{\state'} -
  \TransInd{\sapair}{\state'}) \qvalues(\state', \OptPiOne(\state'))
  \\ &= \sum_{\state'} (\PerTransInd{\sapair}{\state'} -
  \TransInd{\sapair}{\state'}) (\qvalues(\state', \OptPiOne(\state'))
  - (\Trans^\OptPiOne \qvalues)(\sapair)) \\ &=
  \frac{1}{\CompTermEntry(\optsapair) \sqrt{2\Qnumobs}}
  \InvTermInd{\optsapair}{\sapair} \sum_{\state'}
  \TransInd{\sapair}{\state'} (\qvalues(\state', \OptPiOne(\state')) -
  (\Trans^\OptPiOne \qvalues)(\sapair))^2 \\
&= \frac{1}{\CompTermEntry(\optsapair) \sqrt{2\Qnumobs}}
  \InvTermInd{\optsapair}{\sapair} \SigTerm^2(\sapair) \geq 0,
\end{align*}
where the second equality follows from the fact that
$\sum_{\state'} \PerTransInd{\sapair}{\state'} = \sum_{\state'}
\TransInd{\sapair}{\state'} = 1$, the third equality follows by
substituting the value of $\PmatBar$ from
equation~\eqref{eqn:Perturbed-Transition}, and the equality in the
last line follows from the definition~\eqref{eqn:psi-var-defn}.  This
completes the proof of part (c).


\section{Auxiliary lemmas for Theorem~\ref{ThmUpper}}

In this section, we prove the auxiliary lemmas that are used in the
proof of Theorem~\ref{ThmUpper}.

\subsection{Proof of Lemma~\ref{lem:EpochLemma-large}}
\label{Sec:ProofEpochLemma}

This section is devoted to the proof of
Lemma~\ref{lem:EpochLemma-large} which underlies the proof of
Theorem~\ref{ThmUpper}. In order to simplify
notation, we drop the epoch number $\epoch$ from $\Qhat_\epoch$ and
$\Qbar_\epoch$ throughout the remainder of the proof.  Let $\pihat$
and $\pistar$ denote the greedy policies with respect to the $Q$
functions $\Qhat$ and $\Qstar$, respectively. Concretely,
\begin{align}
  \pistar(\state) = \arg\max_{\action \in \actionset} \;\;
  \Qstar(\state, \action) \qquad \pihat(\state) = \arg\max_{\action
    \in \actionset} \;\; \Qhat(\state, \action).
\end{align}
Ties in the $\arg\max$ are broken by choosing the
action~$\action$ with smallest index.

By the optimality of the policies $\pihat$ and $\pistar$ for the
$Q$-functions $\Qhat$ and $\Qstar$, respectively, we have
\begin{gather}
\label{eqn:Qhat-Qstar-fixed-pt-eqn}
  \Qstar = \reward + \contractPar \TranMatQ^{\pistar} \Qstar \quad
  \text{and} \quad \Qhat = \rewardtil + \contractPar
  \TranMatQ^{\pihat} \Qhat, \quad \text{where} \quad
    \rewardtil \defn \reward + \RecenterOp(\Qbar) - \Op(\Qbar).
\end{gather}
In order to complete the proof, we require the following auxiliary
result.
\begin{lemma}
\label{lem:inv-concentration-bound}
We have
\begin{align*}
\inftynorm{(\Id - \contractPar
  \TranMatQ^{\policystar})^{-1}(\rewardtil - \reward)} &\leq
\frac{\inftynorm{\Qbar - \Qstar}}{33} + 4 \cdot
\sqrt{\frac{\LogDM}{\Nm}} \cdot \max_{\policy \in \OptPolicySet}
\inftynorm{\AllCompTerm(\policy; \TranMatQ, \reward)} \\ &\qquad
\qquad + 4 \cdot \frac{\LogDM}{\Nm} \cdot \frac{\VecSpan{\Qstar}}{(1 -
  \contractPar)},
\end{align*}
with probability exceeding $1 - \frac{\pardelta}{8\epochs}$.
\end{lemma}
\noindent See Appendix~\ref{proof:lem:inv-concentration-bound} for the
proof. \\

It remains to prove that under the assumptions of
Lemma~\ref{lem:EpochLemma-large}, the following bound holds with
probability $1 - \frac{\delta}{M}$:
\begin{align}
\label{eqn:instance-ub}
\inftynorm{\Qhat - \Qbar} \leq \inftynorm{(\Id - \contractPar
  \TranMatQ^{\policystar})^{-1}(\rewardtil - \reward)}.
\end{align}
We establish this claim by first showing that the policy $\pihat$ is
an optimal policy, which is achieved in the following lemma.
\begin{lemma}
\label{lem:Pistar-Pihat-Gap}
Given two $Q$-functions $\Qstar$ and $\Qhat$ and the associated
optimal policies $\pistar$ and $\pihat$, we have
\begin{align*}
  \TranMatQ^{\pihat}\Qstar(\state, \action) \geq
  \TranMatQ^{\pistar}\Qstar (\state, \action) - 2\infnorm{\Qhat -
    \Qstar} \quad \text{for all} \quad (\state, \action) \in \stateset
  \times \actionset.
\end{align*}
Moreover, if the batch size satisfies the lower bound $\Nm \geq c_3
\frac{(1 + \infnorm{\reward} + \sigma_r\sqrt{1 - \contractPar})^2}{(1
  - \contractPar)^3} \cdot \frac{\log(\numstates \epochs^2 /
  \pardelta)}{\SpGap^2}$, then $\pihat$ is an optimal policy with
probability at least $1 - \frac{\pardelta}{\epochs}$.  Hence, under the
unique optimal policy (UNQ) condition or Lipschitz (LIP) condition, we have $\TransMatQ^{\pihat} = \TransMatQ^{\pistar}$.
\end{lemma}
We prove this lemma in Appendix~\ref{Pistar-Pihat-gap-proof}. In order
to prove the bound~\eqref{eqn:instance-ub}, it suffices to prove the
following elementwise inequalities:
\begin{subequations}
\begin{align}
\label{eqn:tight-bound-one}
    \Qstar - \Qhat &\preceq (\Id - \contractPar
    \probmat^{\pistar})^{-1}(\reward - \rewardtil) \\
    \label{eqn:tight-bound-two}
    \Qhat - \Qstar &\preceq (\Id - \contractPar
    \probmat^{\pihat})^{-1}(\reward - \rewardtil)
\end{align}
\end{subequations}
Indeed, we have
\begin{align*}
    \abs{\Qstar - \Qhat} \preceq \max\{\abs{\Qstar - \Qhat}_{+},
    \abs{\Qhat - \Qstar}_{+}\} \preceq \abs{(\Id - \contractPar
      \TranMatQ^{\pistar})^{-1}(\reward - \rewardtil)},
\end{align*}
where the maximum operator $\max\{ \cdot, \cdot\}$ is applied
entry-wise. Combining the last two bounds with
Lemma~\ref{lem:Pistar-Pihat-Gap}, and using the lower bound assumption
on the epoch sample size $\Nm$ we obtain
\begin{align*}
  \abs{\Qstar - \Qhat} \preceq \max\{\abs{\Qstar - \Qhat}_{+},
  \abs{\Qhat - \Qstar}_{+}\} &\preceq \abs{(\Id - \contractPar
    \TranMatQ^{\pistar})^{-1}(\reward - \rewardtil), (\Id -
    \contractPar \TranMatQ^{\pihat})^{-1}(\reward - \rewardtil)}
  \notag \\ &\preceq \abs{(\Id - \contractPar
    \TranMatQ^{\pistar})^{-1}(\reward - \rewardtil)},
\end{align*}
where the last inequality uses $\probmatQ^{\pistar} =
\probmatQ^{\pihat}$ (cf. Lemma~\ref{lem:Pistar-Pihat-Gap}). It remains
to prove the bounds~\eqref{eqn:tight-bound-one}
and~\eqref{eqn:tight-bound-two}.

\paragraph{Proof of bounds~\eqref{eqn:tight-bound-one} and~\eqref{eqn:tight-bound-two}:}

By optimality of policies $\pihat$ and $\pistar$ for $Q$-functions
$\Qhat$ and $\Qstar$, respectively, we have
\begin{align}
  \Qstar = \reward + \contractPar \TranMatQ^{\pistar} \Qstar \geqEle
  \reward + \contractPar \TranMatQ^{\pistar} \Qhat \quad \text{and}
  \quad \Qhat = \rewardtil + \contractPar \TranMatQ^{\pihat} \Qhat
  \geqEle \rewardtil + \contractPar \TranMatQ^{\pihat} \Qstar.
\end{align}
This yields:
\begin{align}
\label{eqn:q-gap-one}
  \Qstar - \Qhat &= \reward - \rewardtil +
  \contractPar(\TranMatQ^{\pistar} \Qstar - \TranMatQ^{\pihat} \Qhat)
  \leqEle \reward - \rewardtil + \contractPar\TranMatQ^{\pistar}
  (\Qstar - \Qhat).
\end{align}
Rearranging the last inequality, and using the non-negativity of the
entries of $(\Id - \contractPar\TranMatQ^{\pistar})^{-1}$ we conclude
\begin{align*}
  (\Qstar - \Qhat) \leqEle (\Id -
  \contractPar\TranMatQ^{\pistar})^{-1}(\reward - \rewardtil).
\end{align*}
This completes the proof of the bound~\eqref{eqn:tight-bound-one}. The
proof of bound~\eqref{eqn:tight-bound-two} is similar.

\subsection{Proof of Lemma~\ref{lem:inv-concentration-bound}}
\label{proof:lem:inv-concentration-bound}

Recall the definition \mbox{$\rewardtil \defn \rewardhat +
  \contractPar (\obsmathat^{\pibar} - \TranMatQ^{\pibar}) \Qbar$,}
where $\pibar$ a policy greedy with respect to $\Qbar$; that is,
$\pibar(x) = \arg \max_{u \in \actionset} \Qbar(\state,
\action)$, where we break ties in the $\arg\max$ by choosing the
action~$\action$ with smallest index. We have
\begin{align*}
\inftynorm{(\Id - \contractPar \TranMatQ^{\pistar})^{-1}( \rewardtil -
  \reward)} &\leq \inftynorm{(\Id - \contractPar
  \TranMatQ^{\pistar})^{-1} \left \{ (\rewardhat - \reward) +
  \contractPar(\obsmathat^{\pistar} - \TranMatQ^{\pistar}) \Qstar
  \right\}} \\ &\qquad + \contractPar \inftynorm{ (\Id - \contractPar
  \TranMatQ^{\pistar})^{-1} \left\{ (\obsmathat^{\pibar} \Qbar -
  \obsmathat^{\pistar} \Qstar) - (\TranMatQ^{\pibar} \Qbar -
  \TranMatQ^{\pistar} \Qstar) \right\} }.
\end{align*}

Observe that the random variable $\rewardhat$ and $\obsmathat$ are
averages of $\Nm$ i.i.d.~random variables $\{\NoisyReward_i\}$ and
$\{\obsmathat_i \}$, respectively. Additionally, entrywise, the random
reward is a Gaussian random variable with variance $\sigma_\reward^2$,
and by the generative model assumption, the randomness in the random
rewards $\{ \obsreward_i\}$ is independent of the randomness in $\{
\obsmathat_i \}$. Consequently, applying Hoeffding's bound on the term
involving $\{ \NoisyReward_i \}$, a Bernstein bound on the term
involving $\{ \obsmathat_i \}$ and a union bound yields the following
result which holds with probability at least $1 - \frac{\pardelta}{4
  \epochs}$:
\begin{align*}
&\inftynorm{(\Id - \contractPar \TranMatQ^{\pistar})^{-1} \left \{
    (\rewardhat - \reward) + \contractPar(\obsmathat^{\pistar} -
    \TranMatQ^{\pistar}) \Qstar \right\}} \\ & \qquad \leq
  \frac{4}{\sqrt{\Nm}} \cdot \inftynorm{\AllCompTerm(\pistar;
    \TranMatQ, \reward)} \cdot \sqrt{\LogDM} + \frac{4
    \VecSpan{\Qstar}}{(1 - \contractPar) \Nm} \cdot \LogDM \\ &\qquad
  \leq \frac{4}{\sqrt{\Nm}} \cdot \max_{\policy \in \OptPolicySet}
  \inftynorm{\AllCompTerm(\policy; \TranMatQ, \reward)} \cdot
  \sqrt{\LogDM} + \frac{4 \VecSpan{\Qstar}}{(1 - \contractPar) \Nm}
  \cdot \LogDM .
\end{align*}
Finally, for each state-action pair $(\state, \action)$ the random
variable $(\obsmathat^{\pibar} \Qbar - \obsmathat^{\pistar}
\Qstar)(\state, \action)$ has expectation $(\TranMatQ^{\pibar} \Qbar -
\TranMatQ^{\pistar})(\state, \action)$ with entries uniformly bounded
by $2 \inftynorm{\Qbar - \Qstar}$. Consequently, by a standard
application of Hoeffding's inequality combined with the lower bound
$\Nm \geq c_3 \frac{4^\epoch}{(1 - \contractPar)^2} \LogDM$, we have
\begin{align*}
\frac{\contractPar}{1 - \contractPar} \cdot
\inftynorm{(\obsmathat^{\pibar} \Qbar - \obsmathat^{\pistar} \Qstar) -
  (\TranMatQ^{\pibar} \Qbar - \TranMatQ^{\pistar} \Qstar)} \leq
\frac{\inftynorm{\Qbar - \Qstar}}{33},
\end{align*}
with probability exceeding $1 - \frac{\pardelta}{4\epochs}$. The
statement then follows from combining these two high-probability
statements with a union bound.


\subsection{Proof of Lemma~\ref{lem:EpochLemma-instance-gen}}
\label{Sec:ProofGenEpochLemma}

By Lemma~\ref{lem:gen-SA-bound} and Lemma~\ref{lem:inv-concentration-bound}, it suffices to show
\begin{align*}
\inftynorm{\Qhat_\epoch - \Qstar_\epoch} \leq \frac{1}{2} \inftynorm{(\Id - \contractPar \TranMatQ^{\pistar})^{-1}(\rewardtil - \reward)}
\end{align*}
Recalling the
bounds~\eqref{eqn:tight-bound-one}--\eqref{eqn:tight-bound-two}, we
have
\begin{align*}
\inftynorm{\Qhat_\epoch - \Qstar} &\leq \inftynorm{(\Id -
  \contractPar^{\policystar})^{-1} (\rewardtil - \reward)} +
\contractPar \inftynorm{(\Id - \contractPar
  \TranMatQ^{\policystar})^{-1}(\TranMatQ^{\pihat} -
  \TranMatQ^{\policystar})(\Qhat_\epoch - \Qstar)} \notag \\
& \leq \inftynorm{(\Id - \contractPar
  \TranMatQ^{\pistar})^{-1}(\rewardtil - \reward)} + \frac{L
  \contractPar}{1 - \contractPar} \inftynorm{\Qhat_\epoch - \Qstar}^2
\end{align*}
where the last inequality  uses the Lipschitz
condition~\eqref{EqnLipschitz}.  If we can show ${\frac{L
    \contractPar}{1 - \contractPar} \inftynorm{\Qhat_\epoch - \Qstar}
  \leq \frac{1}{2}}$, we are done. In order to do so, we require the
following auxiliary result:
\begin{lemma}
\label{lem:modified-martin-lem}
Given a batch size $\Nm$ lower bounded as $\Nm \geq c_3
\frac{\LogDM}{(1 - \contractPar)^2}$, we have
\begin{align*}
\inftynorm{\Qhat_\epoch - \Qstar} \leq c_1 \cdot \frac{1 +
  \infnorm{\reward} + \sigma_r\sqrt{1- \contractPar}}{\sqrt{\Nm}(1 -
  \contractPar)^{1.5}} \cdot \LogDMsq
\end{align*}
with probability at least $1 - \frac{\pardelta}{4\epochs}$.
\end{lemma}

With the above lemma at hand and using $\Nm \geq c \LogDM \frac{L^2 (1 + \inftynorm{\reward} + \sigma_\reward\sqrt{1 - \contractPar})^2}{(1 - \contractPar)^5}$ we conclude
\begin{align*}
\inftynorm{\Qhat_\epoch - \Qstar} \leq \frac{1 - \contractPar}{2L},
\end{align*}
as desired. It remains to prove Lemma~\ref{lem:modified-martin-lem}.

\paragraph{Proof of Lemma~\ref{lem:modified-martin-lem}:}

This proof exploits the result of Lemma~\ref{lem:EpochLemma-small}, that with probability at least $1 -
\frac{\pardelta}{\NumEpoch^2}$, we have
\begin{align}
  \label{eqn:bound1}  
  \infnorm{\Qhat_\epoch - \Qstar} \leq \frac{\infnorm{\Qbar_\epoch -
      \Qstar}}{33} + \frac{1 + \infnorm{\reward} + \sigma_\reward
    \sqrt{1 - \contractPar}}{(1 - \contractPar)^{1.5}}
  \sqrt{\frac{\log(8\numstates\epochs^2/\pardelta)}{\Nm}} +
  \frac{\log(8\numstates\epochs^2/\pardelta)}{\Nm} \cdot
  \frac{\VecSpan{\Qstar}}{1 - \contractPar}.
\end{align}
Following an argument similar to the proof of Theorem~\ref{ThmUpper},
we have
\begin{align}
  \infnorm{\Qbar_{\epoch + 1} - \Qstar} & \leq
  \frac{\infnorm{\Qbar_{\epoch} - \Qstar}}{16} + c \left \{ \frac{1 +
    \infnorm{\reward} + \sigma_\reward \sqrt{1 - \contractPar}}{(1 -
    \contractPar)^{1.5}}
  \sqrt{\frac{\log(8\numstates\epochs^2/\pardelta)}{\Nm}} +
  \frac{\log(8\numstates\epochs^2/\pardelta)}{\Nm} \cdot
  \frac{\VecSpan{\Qstar}}{1 - \contractPar} \right \} \notag \\
   & \stackrel{(i)}{\leq} \frac{\infnorm{\Qbar_{1} -
  \Qstar}}{16^{\epoch}} + 2 c \left\lbrace \frac{1 +
  \infnorm{\reward} + \sigma_\reward \sqrt{1 - \contractPar}}{(1 -
  \contractPar)^{1.5}}
  \sqrt{\frac{\log(8\numstates\epochs^2/\pardelta)}{\Nm}} +
  \frac{\log(8\numstates\epochs^2/\pardelta)}{\Nm} \cdot
  \frac{\VecSpan{\Qstar}}{1 - \contractPar} \right\rbrace \notag \\
 &  \stackrel{(ii)}{\leq} \frac{\infnorm{\reward}}{\sqrt{1 -
  \contractPar}}\cdot \frac{1}{(1 - \contractPar)\sqrt{\Nm}} \cdot
  \frac{1}{4^m} \notag \\ \label{eqn:bound2} & \qquad \quad + 2 c
  \left\lbrace \frac{1 + \infnorm{\reward} + \sigma_\reward \sqrt{1 -
  \contractPar}}{(1 - \contractPar)^{1.5}}
  \sqrt{\frac{\log(8\numstates\epochs^2/\pardelta)}{\Nm}} +
  \frac{\log(8\numstates\epochs^2/\pardelta)}{\Nm} \cdot
  \frac{\VecSpan{\Qstar}}{1 - \contractPar} \right\rbrace,
\end{align}
with probability at least $1 - \frac{\pardelta}{4\NumEpoch}$.
Inequality (ii) follows by recursing the first inequality;
the last inequality uses the initialization condition
$\infnorm{\Qbar_1 - \Qstar} \leq \frac{\infnorm{\reward}}{\sqrt{1 -
    \contractPar}}$, and $\Nm \geq \frac{4^m}{(1 - \contractPar)^2}$.
Combining the bounds~\eqref{eqn:bound1} and~\eqref{eqn:bound2} and
using the bounds $\infnorm{\Qstar} \leq \frac{\infnorm{\reward}}{1 -
  \contractPar}$ and $\VecSpan{\Qstar} \leq 2 \inftynorm{\Qstar}$, we
find that
\begin{align*}
\infnorm{\Qhat_\epoch - \Qstar} \leq 8 c \cdot \frac{1 +
  \infnorm{\reward} + \sigma_r \sqrt{1- \contractPar}}{\sqrt{\Nm}(1 -
  \contractPar)^{1.5}} \cdot \log(8 \numstates \epochs^2/\pardelta),
\end{align*}
with probability at least $1 - \frac{\pardelta}{4M}$.  This completes
the proof.


\subsection{Proof of Lemma~\ref{lem:Pistar-Pihat-Gap}} 
\label{Pistar-Pihat-gap-proof}
The proof of this lemma exploits the optimality of the policies
$\pistar$ and $\pihat$ with respect to the $Q$-functions $\Qstar$ and
$\Qhat$, respectively. Accordingly, we have for all state action pair
$(\state, \action) \in \stateset \times \actionset$
\begin{align}
  \TranMatQ^{\pihat} \Qstar(\state, \action) &= \TranMatQ^{\pihat}
  \Qhat(\state, \action) + \TranMatQ^{\pihat} \Qstar(\state, \action)
  - \TranMatQ^{\pihat} \Qhat(\state, \action) \notag \\
& \geq \TranMatQ^{\pistar} \Qhat(\state, \action) - \infnorm{ \Qstar -
    \Qhat} \notag \\
& = \TranMatQ^{\pistar} \Qstar(\state, \action) + \TranMatQ^{\pistar}
  \Qhat(\state, \action) - \TranMatQ^{\pistar} \Qstar(\state, \action)
  - \infnorm{ \Qstar - \Qhat} \notag \\
\label{EqnFinal}
& \geq \TranMatQ^{\pistar} \Qstar(\state, \action) - 2\infnorm{ \Qstar
  - \Qhat}.
\end{align}
The first inequality follows from the optimality of the policy
$\pihat$ with respect to the $Q$-function $\Qhat$.  This completes the
proof of the first part of the lemma.

Turning to the second part, invoking
Lemma~\ref{lem:modified-martin-lem} with a batch size \mbox{$\Nm \geq
  \frac{(1 + \infnorm{\reward} + \sigma_r\sqrt{1 -
      \contractPar})^2}{(1 - \contractPar)^3} \cdot
  \frac{\log(\numstates \epochs^2 / \pardelta)}{\SpGap^2}$} guarantees
that
\begin{align*}
 2 \infnorm{ \Qstar - \Qhat} & < \SpGap.
\end{align*}
This inequality, combined with the bound~\eqref{EqnFinal} and the
definition of the optimality gap $\SpGap$, implies that $\pihat$ is an
optimal policy, and hence $\TransMatQ^{\pihat} = \TransMatQ^{\pistar}$ under the unique policy or Lipschitz assumptions.


\bibliography{bibliography} \bibliographystyle{amsalpha}

\newcommand{\etalchar}[1]{$^{#1}$}
\providecommand{\bysame}{\leavevmode\hbox to3em{\hrulefill}\thinspace}
\providecommand{\MR}{\relax\ifhmode\unskip\space\fi MR }
\providecommand{\MRhref}[2]{%
  \href{http://www.ams.org/mathscinet-getitem?mr=#1}{#2}
}
\providecommand{\href}[2]{#2}
\begin{thebibliography}{SWW{\etalchar{+}}18}

\bibitem[AMK13]{Azar2013Minimax}
Mohammad~Gheshlaghi Azar, R{\'e}mi Munos, and Hilbert~J Kappen, \emph{Minimax
  {PAC} bounds on the sample complexity of reinforcement learning with a
  generative model}, Machine Learning \textbf{91} (2013), no.~3, 325--349.

\bibitem[Ber09]{Bertsekas2009}
Dimitri~P. Bertsekas, \emph{Neuro-dynamic programming}, Springer US, Boston,
  MA, 2009.

\bibitem[Bir83]{birge1983}
Lucien Birg{\'e}, \emph{Approximation dans les espaces m{\'e}triques et
  th{\'e}orie de l'estimation}, Zeitschrift f{\"u}r Wahrscheinlichkeitstheorie
  und Verwandte Gebiete \textbf{65} (1983), no.~2, 181--237.

\bibitem[BRS18]{pmlr-v75-bhandari18a}
Jalaj Bhandari, Daniel Russo, and Raghav Singal, \emph{A finite time analysis
  of temporal difference learning with linear function approximation},
  Conference On Learning Theory, PMLR, 2018, pp.~1691--1692.

\bibitem[CL15]{cailow2015estconv}
T~Cai and Mark Low, \emph{A framework for estimation of convex functions},
  Statistica Sinica \textbf{25} (2015), 423--456.

\bibitem[DSTM18]{dalal2018finite}
Gal Dalal, Bal{\'a}zs Sz{\"o}r{\'e}nyi, Gugan Thoppe, and Shie Mannor,
  \emph{Finite sample analyses for {TD} (0) with function approximation}, AAAI
  Conference on Artificial Intelligence, vol.~32, 2018, pp.~6144--6153.

\bibitem[JJS94]{Jaakkola1994@StochDP}
Tommi Jaakkola, Michael~I Jordan, and Satinder~P Singh, \emph{On the
  convergence of stochastic iterative dynamic programming algorithms}, Neural
  Computation \textbf{6} (1994), no.~6, 1185--1201.

\bibitem[JZ13]{johnson2013accelerating}
Rie Johnson and Tong Zhang, \emph{Accelerating stochastic gradient descent
  using predictive variance reduction}, Advances in Neural Information
  Processing Systems, vol.~26, 2013, pp.~315--323.

\bibitem[KPR{\etalchar{+}}20]{khamaru2020PE}
Koulik Khamaru, Ashwin Pananjady, Feng Ruan, Martin~J Wainwright, and Michael~I
  Jordan, \emph{Is temporal difference learning optimal? an instance-dependent
  analysis}, arXiv preprint arXiv:2003.07337 (2020), 1--38.

\bibitem[LFDA16]{JMLR:v17:15-522}
Sergey Levine, Chelsea Finn, Trevor Darrell, and Pieter Abbeel,
  \emph{End-to-end training of deep visuomotor policies}, Journal of Machine
  Learning Research \textbf{17} (2016), no.~1, 1334--1373.

\bibitem[LS18]{pmlr-v84-lakshminarayanan18a}
Chandrashekar Lakshminarayanan and Csaba Szepesvari, \emph{Linear stochastic
  approximation: How far does constant step-size and iterate averaging go?},
  AISTATS: Conference on AI and Statistics, vol.~21, PMLR, 2018,
  pp.~1347--1355.

\bibitem[MMM14]{maillard2014hard}
Odalric-Ambrym Maillard, Timothy~A Mann, and Shie Mannor, \emph{How hard is my
  {MDP}? "{T}he distribution-norm to the rescue"}, Advances in Neural
  Information Processing Systems, vol.~27, 2014, pp.~1835--1843.

\bibitem[PB79]{putermanbrumelle1979}
Martin~L. Puterman and Shelby~L. Brumelle, \emph{On the convergence of policy
  iteration in stationary dynamic programming}, Mathematics of Operations
  Research \textbf{4} (1979), no.~1, 60--69.

\bibitem[Put14]{puterman2014markov}
Martin~L Puterman, \emph{{M}arkov {D}ecision {P}rocesses: Discrete stochastic
  dynamic programming}, John Wiley \& Sons, 2014.

\bibitem[PW20]{pananjady2020instancedependent}
A.~{Pananjady} and M.~J. {Wainwright}, \emph{Instance-dependent
  $\ell_\infty$-bounds for policy evaluation in tabular reinforcement
  learning}, IEEE Transactions on Information Theory \textbf{67} (2020), no.~1,
  566--585.

\bibitem[SB18]{sutton1998}
Richard~S. Sutton and Andrew~G. Barto, \emph{Reinforcement learning: An
  introduction}, second ed., The MIT Press, 2018.

\bibitem[SHM{\etalchar{+}}16]{silver2016alphago}
David Silver, Aja Huang, Chris~J Maddison, Arthur Guez, Laurent Sifre, George
  Van Den~Driessche, Julian Schrittwieser, Ioannis Antonoglou, Veda
  Panneershelvam, and Marc Lanctot, \emph{Mastering the game of {G}o with deep
  neural networks and tree search}, Nature \textbf{529} (2016), no.~7587,
  484--489.

\bibitem[SJ19]{simchowitz2019non}
Max Simchowitz and Kevin Jamieson, \emph{Non-asymptotic gap-dependent regret
  bounds for tabular {MDP}s}, Advances in Neural Information Processing
  Systems, vol.~33, 2019, pp.~1153--1162.

\bibitem[SWW{\etalchar{+}}18]{sidford2018near}
Aaron Sidford, Mengdi Wang, Xian Wu, Lin~F Yang, and Yinyu Ye,
  \emph{Near-optimal time and sample complexities for solving markov decision
  processes with a generative model}, Advances in Neural Information Processing
  Systems, vol.~33, 2018, pp.~5192--5202.

\bibitem[SWWY18]{sidford2018variance}
Aaron Sidford, Mengdi Wang, Xian Wu, and Yinyu Ye, \emph{Variance reduced value
  iteration and faster algorithms for solving markov decision processes},
  ACM-SIAM Symposium on Discrete Algorithms, vol.~29, SIAM, 2018, pp.~770--787.

\bibitem[Sze97]{szepesvari1997asymptotic}
Csaba Szepesv{\'a}ri, \emph{The asymptotic convergence-rate of {$Q$}-learning},
  Advances in Neural Information Processing Systems, vol.~10, 1997,
  pp.~1064--1070.

\bibitem[TFR{\etalchar{+}}17]{tobin2017domain}
Josh Tobin, Rachel Fong, Alex Ray, Jonas Schneider, Wojciech Zaremba, and
  Pieter Abbeel, \emph{Domain randomization for transferring deep neural
  networks from simulation to the real world}, International Conference on
  Intelligent Robots and Systems (IROS), IEEE, 2017, pp.~23--30.

\bibitem[Tsi94]{tsitsiklis1994}
John~N Tsitsiklis, \emph{Asynchronous stochastic approximation and
  {$Q$}-learning}, Machine Learning \textbf{16} (1994), no.~3, 185--202.

\bibitem[Wai19a]{dinosaur2019}
Martin~J. Wainwright, \emph{High-dimensional statistics: A non-asymptotic
  viewpoint}, Cambridge Series in Statistical and Probabilistic Mathematics,
  Cambridge University Press, 2019.

\bibitem[Wai19b]{wainwright2019stochastic}
Martin~J Wainwright, \emph{Stochastic approximation with cone-contractive
  operators: Sharp $\ell_\infty$-bounds for {$Q$}-learning}, Tech. report,
  2019, arXiv preprint arXiv:1905.06265.

\bibitem[Wai19c]{wainwright2019variancereduced}
\bysame, \emph{Variance-reduced {$Q$}-learning is minimax optimal}, Tech.
  report, 2019, arXiv preprint arXiv:1906.04697.

\bibitem[WD92]{Watkins1992Qlearning}
Christopher~JCH Watkins and Peter Dayan, \emph{{$Q$}-learning}, Machine
  Learning \textbf{8} (1992), no.~3-4, 279--292.

\bibitem[ZB19]{zanette2019tighter}
Andrea Zanette and Emma Brunskill, \emph{Tighter problem-dependent regret
  bounds in reinforcement learning without domain knowledge using value
  function bounds}, International Conference on Machine Learning, PMLR, 2019,
  pp.~7304--7312.

\bibitem[ZKB19]{zanette2019almost}
Andrea Zanette, Mykel~J Kochenderfer, and Emma Brunskill, \emph{Almost
  horizon-free structure-aware best policy identification with a generative
  model}, Advances in Neural Information Processing Systems, vol.~32, 2019,
  pp.~5625--5634.

\end{thebibliography}


\end{document}